\documentclass{article}

\usepackage[final]{styles}

\usepackage{amsmath,amsfonts,bm}

\def\eqref#1{equation~\ref{#1}}

\def\1{\bm{1}}

\def\rvtheta{{\mathbf{\theta}}}

\def\rvw{{\mathbf{w}}}
\def\rvx{{\mathbf{x}}}
\def\rvy{{\mathbf{y}}}
\def\rvz{{\mathbf{z}}}

\DeclareMathAlphabet{\mathsfit}{\encodingdefault}{\sfdefault}{m}{sl}
\SetMathAlphabet{\mathsfit}{bold}{\encodingdefault}{\sfdefault}{bx}{n}

\def\gB{{\mathcal{B}}}

\def\gE{{\mathcal{E}}}
\def\gF{{\mathcal{F}}}
\def\gG{{\mathcal{G}}}
\def\gH{{\mathcal{H}}}

\def\gK{{\mathcal{K}}}

\def\gN{{\mathcal{N}}}

\def\gQ{{\mathcal{Q}}}

\def\gT{{\mathcal{T}}}

\def\gV{{\mathcal{V}}}

\def\sR{{\mathbb{R}}}

\newcommand{\KL}{{\mathrm{KL}}}

\DeclareMathOperator*{\argmin}{argmin}

\def\pd{p_{\text{data}}}

\def\dd{\mathrm{d}}

\def\dtheta{\nabla_{\rvtheta}}
\def\gradx{\nabla_{\rvx}}
\def\ratio{r(\rvx_{\theta}; \theta)}
\def\ratios{r(\rvx_{\theta}; \theta_s)}
\def\ratiov{r(\rvx;\theta)}

\def\pz{\xi}

\def\wgrad{\nabla_{W_2}}

\def\ee{\mathbb{E}}
\def\fkl{\mathcal{F}_{\text{kl}}}
\def\eez{\mathbb{E}_{\rvz \sim \pz}}
\def\epx{\mathbb{E}_{\rvx \sim p}}
\def\eqx{\mathbb{E}_{\rvx \sim q_\theta}}
\def\eqt{\mathbb{E}_{\rvx \sim q_t}}
\def\setS{\mathbb{S}^{n\times n}}
\def\setC{\mathbb{C}^{n\times n}}
\def\ddpi{\dd \pi}

\usepackage[utf8]{inputenc} 
\usepackage[T1]{fontenc}    

\usepackage[toc,page,header]{appendix}
\usepackage{minitoc}

\title{Bridging the Gap Between Variational Inference and Wasserstein Gradient Flows}

\author{%
  Mingxuan Yi, \quad Song Liu\\
  School of Mathematics\\
  University of Bristol\\
  \texttt{\{mingxuan.yi, song.liu\}@bristol.ac.uk} \\
}

\begin{document}
\doparttoc 
\faketableofcontents 

\maketitle

\begin{abstract}
Variational inference is a technique that approximates a target distribution by optimizing within the parameter space of variational families. On the other hand, Wasserstein gradient flows describe optimization within the space of probability measures where they do not necessarily admit a parametric density function.
In this paper, we bridge the gap between these two methods. We demonstrate that, under certain conditions, the Bures-Wasserstein gradient flow can be recast as the Euclidean gradient flow where its forward Euler scheme is the standard black-box variational inference algorithm. Specifically, the vector field of the gradient flow is generated via the path-derivative gradient estimator. We also offer an alternative perspective on the path-derivative gradient, framing it as a distillation procedure to the Wasserstein gradient flow. Distillations can be extended to encompass $f$-divergences and non-Gaussian variational families. This extension yields a new gradient estimator for $f$-divergences, readily implementable using contemporary machine learning libraries like PyTorch or TensorFlow.
\end{abstract}
\section{Introduction}
An inference problem is generally difficult because it often requires dealing with a probability distribution only known up to a normalizing constant. Traditional statistical methods, such as Markov Chain Monte Carlo (MCMC), provide approximate solutions to such problems. However, MCMC struggles with high-dimensional challenges and is computationally intensive. An alternative is variational inference  \citep{jordan1999introduction, blei2017variational}, an optimization-based method to approximate the target probability distribution with a member (variational distribution) from a family of parametric models, denoted as $\{q(\rvx;\theta)| \theta \in \Theta\}$. Variational inference achieves these approximations by minimizing statistical divergences, which measure the disparity between the variational distribution $q(\rvx;\theta)$ and the target distribution $p(\rvx)$. For example, a commonly used measurement is the (reverse) Kullback-Leibler (KL) divergence. Using this measurement, variational inference finds the best $q(\rvx;\hat{\theta})$ via,
\begin{equation*}
\hat{\theta} = \argmin_{\theta \in \Theta}\KL(q_\theta||p) = \argmin_{\theta \in \Theta}\mathbb{E}_{\rvx \sim q_\theta} \left[ \log \frac{q(\rvx;\theta)}{p(\rvx)} \right] \footnote{{$q_\theta$} represents $q(\rvx;\theta)$.}.
\end{equation*}

The advantage of variational inference is its adaptability. It can be applied to a wide range of models, from classical Bayesian models to more complex deep generative models. Furthermore, with the rise of deep learning libraries like TensorFlow \citep{abadi2016tensorflow} and PyTorch \citep{paszke2019pytorch}, variational inference can be easily implemented, scaled, and integrated with neural networks, making it a popular choice for modern machine learning tasks. Most VI techniques hinge on deriving particular evidence bounds, facilitating the acquisition of a gradient estimator suitable for optimization.

Wasserstein gradient flows \citep{ambrosio2005gradient} characterize a particle flow differential equation in the sample space where its associated marginal probability evolves with time to decrease a functional, e.g., the KL divergence as well. This differs from variational inference because the functional is decreased over the whole space of probability distributions where the distribution does not necessarily admit a parametric density function. A typical example is the Langevin stochastic differential equation (SDE),
\begin{equation} \label{langevin_sde}
     \dd \rvx_t =\gradx \log p(\rvx_t) \dd t + \sqrt{2} \dd \rvw_t,
\end{equation}
where $\rvw_t$ is the standard Wiener process, its marginals $\{q_t\}_{t\geq0}$ can be viewed as the Wasserstein gradient flow of the KL divergence \citep{jordan1998variational, otto2001geometry}. Wasserstein gradient flows have been widely studied in deep generative modelings \citep{ansari2020refining, glaser2021kale, yi2023monoflow} and sampling methods \citep{bernton2018langevin, cheng2018convergence, wibisono2018sampling, chewi2020svgd}. A previous work \citep{lambert2022variational} links Wasserstein gradient flows to variational inference by assuming the marginal probability admits a parametric Gassusian density function such that the continuous evolution of marginals and its discretization scheme can be obtained under the Bures-Wasserstein geometry. 
 
Both variational inference and Wasserstein gradient flows optimize probability distributions by minimizing certain statistical discrepancies. However, they appear to operate in parallel rather than intersecting domains. In this paper, we bridge the gap between variational inference and Wasserstein gradient flows. Specifically, we unveil a surprising result that the Bures-Wasserstein gradient flow \citep{lambert2022variational} can be translated into a Euclidean gradient flow where its forward Euler scheme is exactly the black-box variational inference (BBVI) algorithm, with the Gaussian family but under different parameterizations. We show that the ordinary differential equation (ODE) system describing the Bures-Wasserstein gradient flow can be obtained via the path-derivative gradient estimator \citep{roeder2017sticking}. We further establish that the connection between the Euclidean gradient flow and the Bures-Wasserstein gradient flow arises from Riemannian submersion. In addition, we provide an alternative view on the path-derivative gradient as distillation which can be generalized to general $f$-divergences and non-Gaussian variational families. We obtain a novel gradient estimator for $f$-divergences which can be implemented by Pytorch or TensorFlow. We summarize our contributions as:

\begin{enumerate}
    \item We bridge the gap between black-box variational inference and the Bures-Wasserstein gradient flows by showing the equivalence between them under certain conditions.
    \item We provide an insight into the geometry on BBVI. This illustrates that the standard BBVI minimizing the KL divergence in the Euclidean geometry naturally involves the Wasserstein geometry.
    \item We propose an alternative implementation of the path-derivative gradient estimator which can be generalized to $f$-divergences and non-Gaussian families.
    \item A novel unbiased path-derivative gradient estimator of $f$-divergences is derived and this estimator generalizes previous works. 

\end{enumerate}
\section{Background}
In this section, we review preliminaries on variational inference and Wasserstein gradient flows. 
In this paper, all probability distributions are defined over the sample space $\mathbb{R}^n$, i.e., the standard Euclidean space. 

\subsection{Variational Inference}
Variational inference (VI) reformulates inference problems as optimization problems.  To allow for the optimization via the gradient descent algorithm, we need to compute the gradient of the KL divergence with respect to the parameter $\theta$.
If there exists a reparameterization $\rvx_\theta = g(\rvz;\theta) \sim q_\theta$, $\rvz \sim \pz$ where $\pz$ is a base distribution and $g$ transforms $\rvz$ to $\rvx_\theta$, we can obtain the gradient of the KL divergence as 
\begin{equation}
     \dtheta \KL(q_\theta||p) =  \mathbb{E}_{\rvz \sim \pz} \left[\dtheta \log \frac{q(\rvx_\theta;\theta)}{p(\rvx_\theta)} \right] \label{rep_grad}.
\end{equation}
The above \myeqref{rep_grad} is called the reparameterization gradient estimator \citep{kingma2013auto, rezende2014stochastic}. The reparameterization gradient can be effortlessly implemented using auto-differentiation tools such as PyTorch \citep{paszke2019pytorch} and TensorFlow \citep{abadi2016tensorflow}, allowing us to bypass the need for model-specific derivations and enabling us to perform variational inference in a black-box manner \citep{ranganath2014black}, see \autoref{alg}. An alternative is to use the score function gradient, details regarding the score function gradient can be found in \citep{mohamed2019monte}.
\begin{algorithm}
\caption{Black-Box Variational Inference (Reparameterization gradient)} \label{alg}
\begin{algorithmic}
\Require target distribution $p(\rvx)$, variational distribution $q(\rvx;\theta)$, learning rate $\tau$.
\While{not converged}
 \State 1. Sample $\rvx^i_\theta = g(\rvz^i;\theta) \sim q_\theta$, $\rvz^i \sim \pz$.
 \State 2. Evaluate $L(\theta)=  \frac{1}{N}\sum_i \log \big[ q(\rvx^i_\theta; \theta)/p(\rvx^i_\theta)\big]$.
 \State 3. $\theta \leftarrow \theta - \tau \dtheta L(\theta)$ via back-propagation.
 \EndWhile
\end{algorithmic}
\end{algorithm}

However, the target distribution is sometimes only represented by an unnormalized density function $p(\rvx)$. In order to evaluate the true density, we need to evaluate the normalizing constant $C = \int p(\rvx)\dd \rvx$ such that $p_{\text{true}}(\rvx) = p (\rvx) / C $, which is generally intractable. For example, to obtain the true density of the posterior distribution in Bayesian inference, the necessity arises to normalize the product of the likelihood and the prior, a task that frequently entails dealing with intractable integration. Variational inference mitigates this issue by leveraging the linearity of the logarithm function such that the normalizing constant does not affect the minimization of the KL divergence since
\begin{equation*}
    \KL(q_\theta||p) = \KL(q_\theta||p_{\text{true}})  +\log C.
\end{equation*}
$-\KL(q_\theta||p)$ is called the evidence lower bound (ELBO) in the Bayesian inference setting \citep{blei2017variational}.

\subsection{Wasserstein Gradient Flows}
Wasserstein gradient flows formulate the evolution of probability distributions over time by decreasing a functional on $\mathcal{P}(\mathbb{R}^n)$, where $\mathcal{P}(\mathbb{R}^n)$ refers to the space of probability distribution over $\mathbb{R}^n$ with finite second moments.
Let $(\mathcal{P}(\mathbb{R}^n), W_2)$ be a metric space of $\mathcal{P}(\mathbb{R}^n)$ equipped with Wasserstein-2 distance, and we denote this space as Wasserstein space. A curve $\{q_t\}_{t\geq 0}\in \mathcal{P}(\mathbb{R}^n)$ in the Wasserstein space is said to be the gradient flow of a functional $\mathcal{F} \colon \mathcal{P}(\mathbb{R}^n) \to \mathbb{R}$ if it satisfies the following continuity equation \citep{ambrosio2005gradient},
\begin{equation}
    \frac{\partial {q_t}}{\partial t} = \text{div}\big(q_t \nabla_{W_2} {\mathcal{F}(q_t)}\big). \label{wgf}
\end{equation}
$\nabla_{W_2} {\mathcal{F}(q)}$ is called the Wasserstein gradient of the functional $\mathcal{F}(q)$ which satisfies
\begin{equation*}
    \nabla_{W_2} {\mathcal{F}(q)} =  \gradx {\frac{\delta \mathcal{F}(q)}{\delta q}},
\end{equation*}
where $\gradx$ is the Euclidean gradient operator and $\delta \mathcal{F}(q) / \delta q$ is the first variation of $\mathcal{F}(q)$. The Wasserstein gradient defines a family of vector fields $\{v_t\}_{t \geq 0}$ in Euclidean space $\mathbb{R}^n$ which characterizes a probability flow ordinary differential equation (ODE), 
\begin{equation}
    \dd \rvx_t = v_t(\rvx_t) \dd t = - \nabla_{W_2} {\mathcal{F}(q_t)} (\rvx_t) \dd t. \label{ode_vec}
\end{equation}
This ODE describes the evolution of particle $\rvx_t\sim q_t$ in $\mathbb{R}^n$ where the associated marginal $q_t$ evolves to decrease $\mathcal{F}(q_t)$ along the direction of steepest descent according to the continuity equation in \myeqref{wgf}. 

The Wasserstein gradient flow can be discretized via the following movement minimization scheme with step size $\tau$, also known as the Jordan-Kinderlehrer-Otto (JKO) scheme\footnote{$q^{\tau}_k$ denotes the discretization of $q_t$ with step size $\tau$ where $k$ is the index of the discretized time.} \citep{jordan1998variational},
\begin{equation}
    q^{\tau}_{k+1} = \argmin_{q \in \mathcal{P}(\mathbb{R}^n)} \left\{\mathcal{F}(q) + \frac{1}{2\tau} W^2_2 (q,q^{\tau}_{k})\right\}, \label{jko}
\end{equation}
the JKO scheme is to encourage $q^{\tau}_{k+1}$ to minimize the functional $\mathcal{F}(q)$ but stay close to $q^{\tau}_{k}$ in Wasserstein-2 distance as much as possible. It can be shown that as $\tau \to 0$, the limiting solution of \myeqref{jko} coincides with the curve $\{q_t\}_{t\geq 0}$ defined by the continuity equation in \myeqref{wgf}.

A special case of Wasserstein gradient flows is under the KL divergence $\fkl(q) = \KL(q||p)$, the continuity equation reads the Fokker-Planck equation
\begin{equation*}
     \frac{\partial {q_t}}{\partial t} = \text{div}\big(q_t (\gradx \log q_t- \gradx\log p)\big),
\end{equation*}
where the Wasserstein gradient is $\wgrad \fkl(q_t) = \gradx \log (q_t/p)$ and the probability flow ODE follows
\begin{equation}
     \dd \rvx_t =\big(\gradx \log p(\rvx_t)- \gradx\log q_t(\rvx_t)\big) \dd t.\label{prob_ode}
\end{equation}
The Fokker-Planck equation is also the continuity equation of the Langevin SDE in \myeqref{langevin_sde}. The Langevin SDE and the probability flow ODE share the same marginals $\{q_t\}_{t\geq 0}$ if they evolve from the same initial $q_0$.


\section{Bures-Wasserstein Gradient Flows}
In this section, we briefly review gradient flows defined in the Bures-Wasserstein space $(\mathcal{BW}{(\mathbb{R}^n)}, W_2)$, i.e., the subspace of the Wasserstein space consisting of Gaussian distributions. We further show that black-box variational inference (BBVI) with the Gaussian family realizes the forward Euler scheme to the Bures-Wasserstein Gradient flows. Specifically, the vector fields of the ODE system describing the evolution of Gaussian mean and covariance \citep{lambert2022variational} can be obtained by the path-derivative (sticking the landing) gradient estimator \citep{roeder2017sticking}.

\subsection{Bures-Wasserstein JKO Scheme}\label{3.1}
Recall that the Wasserstein-2 distance between two Gaussian distributions $q=\mathcal{N}(\mu, \Sigma)$ and $p=\mathcal{N}(\mu_p, \Sigma_p)$ has a closed form,
\begin{equation} \label{w_2_gauss}
    W_2^2 (q, p) = \Vert \mu - \mu_p \Vert_2^2 + \mathcal{B}^2(\Sigma, \Sigma_p),
\end{equation}
where $\mathcal{B}^2(\Sigma, \Sigma_p)=\text{tr}(\Sigma+\Sigma_p-2(\Sigma^{\frac{1}{2}}\Sigma_p\Sigma^{\frac{1}{2}})^{\frac{1}{2}})$ is the squared Bures distance \citep{bures1969extension}. 
By restricting the JKO scheme to the Bures-Wasserstein space, 
\begin{equation*}
    q^{\tau}_{k+1} = \argmin_{q \in \mathcal{BW}(\mathbb{R}^n)} \left\{\KL(q||p) + \frac{1}{2\tau} W^2_2 (q,q^{\tau}_{k})\right\}, \label{bures_jko}
\end{equation*}
\cite{lambert2022variational} showed that the above discretization scheme yields a limiting curve $\{q_t:\mathcal{N}(\mu_t, \Sigma_t)\}_{t\geq 0}$ as a gradient flow of the KL divergence in the Bures-Wasserstein space where the means and covariance matrices of Gaussians follow an ODE system,
\begin{equation}
\begin{split}
    &\frac{\dd \mu_t}{\dd t} =  \mathbb{E}_{\rvx \sim q_t}\left[ \gradx \log \frac{p(\rvx)}{ q_t(\rvx)}\right],\\
    &\frac{\dd \Sigma_t}{\dd t} =   \mathbb{E}_{\rvx \sim q_t}\left[\left(\gradx \log \frac{p(\rvx)}{ q_t(\rvx)} \right)^T(\rvx - \mu_t) \right] +\mathbb{E}_{\rvx \sim q_t}\left[ (\rvx-\mu_t)^T \gradx \log \frac{p(\rvx)}{ q_t(\rvx)} \right].\label{ode_system}
\end{split}
\end{equation}
{\color{red}Notice that the Gaussian mean $\mu$ in this paper is {{the row vector}}}, while some other literature uses the column vector formulation.

 \textbf{Remark 1.} \myeqref{ode_system} the Hessian-free form of the covariance evolution, see Appendix A in \citep{lambert2022variational} or Appendix \ref{equal_ode} in this paper for other equivalent forms of the ODE system, such as the Sarkka equation \citep{sarkka2007unscented} in Bayesian filtering.

\subsection{Unrolling Black-Box Variational Inference}
In this section, we will show how black-box variational inference (BBVI) leads to the same ODE in Eq.~(\ref{ode_system}). 
Performing BBVI with the standard gradient descent algorithm with the learning rate $\tau$ follows the iteration 
\begin{equation}
\theta^\tau_{k+1} = \theta^\tau_k - \tau \nabla_\theta \KL(q_\theta||p)\big\vert_{\theta=\theta^\tau_k}.
    \label{gd}
\end{equation}
We examine a particular kind of gradient estimator for the KL divergence as detailed below.
Given the reparameterization $\rvx_\theta = g(\rvz;\theta) \sim q_\theta$, $\rvz \sim \pz$,  the reparameterization gradient in \myeqref{rep_grad} can be decomposed into two terms \citep{roeder2017sticking},
\begin{equation}
\begin{split}
    & \dtheta \KL(q_\theta||p) =  \mathbb{E}_{\rvz \sim \pz} \left[\dtheta \log \frac{q(\rvx_\theta;\theta)}{p(\rvx_\theta)} \right]=   \color{blue}\underbrace{\color{black}\mathbb{E}_{\rvz \sim \pz} \left[ \gradx \log \frac{q(\rvx_\theta;\theta)}{p(\rvx_\theta)} \circ \dtheta \rvx_\theta \right]}_\text{path-derivative gradient}  \color{black}+\color{red}\underbrace{\color{black}\bcancel{\mathbb{E}_{\rvx\sim q_\theta} \left[ \dtheta \log q(\rvx;\theta) \right].}}_\text{score function}
\end{split}\label{stl}
\end{equation}
\textbf{Remark 2.} The symbol "$\circ$" represents the application of the chain rule for each element in $\theta$ during backpropagation, e.g., if $\theta$ is a list $(\mu, S)$, we apply the chain rule to $\mu$ and $S$ individually. Such manipulation can be simply implemented via auto-differentiation libraries, we refer to \citep{paszke2019pytorch} and 
\citep{abadi2016tensorflow} for more details.

The first term in \myeqref{stl} is called path-derivative gradient \citep{mohamed2019monte} since it requires differentiation through the reparameterized variable $\rvx_\theta$ which encodes the pathway from $\theta$ to the KL divergence. The second term cancels out because the score function has a zero mean. \cite{roeder2017sticking} proposed a simple efficient approach to implement this path-derivative gradient via a stop gradient operator, e.g., ``\texttt{detach}'' in PyTorch or ``\texttt{stop\_gradient}'' in TensorFlow. We use the notation $\theta_s$ to denote the application of the stop gradient operator to the parameter $\theta$. Once such an operator is applied, the differentiation through the variational parameter $\theta$ is discarded, i.e., $\theta_s$ can be regarded as a constant which is no longer trainable. 
Therefore, we can write the path-derivative gradient as 
\begin{equation}
\dtheta \KL(q_\theta||p)=     \mathbb{E}_{\rvz \sim \pz} \left[ \dtheta \log \frac{q(\rvx_\theta;\theta_s)}{p(\rvx_\theta)} \right]. \label{path}
\end{equation}
More details on the path-derivative gradient can be found in Appendix \ref{path app}. 

Unlike the previous Section \ref{3.1}, we now consider the Gaussian variational family $\mathcal{N}(\mu, SS^T)$ with the parameter $\theta=(\mu, S)$ where $S$ is the scale matrix, which avoids matrix decomposition to allow for the efficient reparameterization $\rvx_\theta = \mu + \rvz S^T \in \mathcal{N}(\mu, SS^T)$ with $ \rvz \in \mathcal{N}(0, I).$ \autoref{rule} provides a specific expression for the path derivative gradient given this Gaussian variational family. 

\begin{proposition}
\label{rule}
If $q(\rvx;\theta)=\mathcal{N}(\mu, \Sigma)$ with $\Sigma=SS^T$ is a  Gaussian distribution with parameter $\theta = (\mu, S)$ and the reparameterization is given by $\rvx_\theta = \mu + \rvz S^T, \rvz \sim \mathcal{N}(0, I)$.  The path-derivative gradient in \myeqref{stl} or \myeqref{path} is given by
\begin{equation}
\begin{split}
    &\nabla_\mu \KL(q_\theta||p) =- \mathbb{E}_{\rvx \sim q_\theta} \left[ \gradx \log \frac{p(\rvx)}{ q(\rvx;\theta)} \right] ,\\
    &\nabla_S \KL(q_\theta||p) =   -\mathbb{E}_{\rvx \sim q_\theta}\left[\left(\gradx \log \frac{p(\rvx)}{ q(\rvx;\theta)} \right)^T(\rvx - \mu) S ^{-T}\right].\label{euc_grad}
\end{split}
\end{equation}
\end{proposition}
See the proof of \autoref{rule} in Appendix \ref{proof_p1}.
By letting $\tau \to 0$, the gradient descent algorithm in \myeqref{gd} corresponds to an ODE system for $\theta_t = (\mu_t, S_t)$, 
\begin{equation}
\begin{split}
    \frac{\dd \mu_t}{\dd t} &=  \mathbb{E}_{\rvx \sim q_t}\left[ \gradx \log \frac{p(\rvx)}{ q_t(\rvx)}\right],\\
    \frac{\dd S_t}{\dd t} &=   \mathbb{E}_{\rvx \sim q_t}\left[\left(\gradx \log \frac{p(\rvx)}{ q_t(\rvx)} \right)^T (\rvx - \mu_t)  S_t ^{-T}\right].\label{scale_ode}
\end{split}
\end{equation}
Using the fact $\dd \Sigma_t = (\dd S_t) S_t^T + S_t (\dd S_t^{T})$, \myeqref{scale_ode} implies the covariance evolution in Eq.~(\ref{ode_system}). Note that the converse is not true because given a covariance matrix $\Sigma$, its decomposition $\Sigma=SS^T$ is not unique. 

\autoref{rule} suggests that the Bures-Wasserstein gradient flow can be equivalently derived through an alternative parameterization of Gaussians. Performing BBVI using the standard gradient descent (no momentum) is exactly the forward Euler scheme to the ODE in \myeqref{scale_ode}, although we still need to evaluate gradient estimators via Monte Carlo methods.

\subsubsection{Geometry on Black-Box Variational Inference}
The previous result is surprising because we have not introduced any specific geometry to BBVI but it leads to the same ODE system which is derived from the Bures-Wasserstein geometry. In this section, we advance our discussion by providing a comprehensive geometric analysis on BBVI.

Given $q^{\tau}_{k} = \mathcal{N}(\mu^{\tau}_k, S^{\tau}_k {S^{\tau}_k}^T)$, consider the following discretization scheme, 
\begin{equation}
    q^{\tau}_{k+1} = \argmin_{(\mu, S)\in \Theta} \Big\{\KL(q_\theta||p) + \frac{1}{2\tau} \big(\Vert\mu -\mu^{\tau}_k \Vert_2^2 +\Vert S -S^{\tau}_k \Vert_{\text{F}}^2 \big) \Big\}, \label{proximal}
\end{equation}
where $\Theta: \mathbb{R}^n \times \setS$ is the parameter space and $\setS$ denotes the space of non-singular matrices. $\Vert A -B \Vert_{\text{F}}^2=\mathrm{tr}(AA^T+BB^T-2A^TB)$ is the squared Frobenius distance and its derivative is given by $\nabla_{A}\Vert A -B \Vert_{\text{F}}^2 = 2(A - B)$. Therefore, similar to the proximal method in Euclidean space, the discretization scheme in \myeqref{proximal} leads to an implicit iteration for $\theta^\tau_{k} = (\mu^{\tau}_k, S^{\tau}_k)$, 
\begin{equation*}
\theta^\tau_{k+1} = \theta^\tau_k - \tau \nabla_\theta \KL(q_\theta||p)\big\vert_{\theta=\theta^\tau_{k+1}}.
\end{equation*}
which is the backward Euler scheme to the ODE in \myeqref{scale_ode}. It is obvious that the Frobenius distance between scales is equal to the Bures distance 
    in \myeqref{w_2_gauss} between covariances if the variational family is a mean-field Gaussian (diagonal covariance), but this is not true for the general case. 

Similar to \cite{takatsu2011wasserstein}, next we only focus on the covariances of Gaussians by assuming they have zero means, since the difference between means in both Eq.~(\ref{proximal}) and Eq.~(\ref{w_2_gauss}) is just the Euclidean distance, which can be trivially generalized afterward.
We consider two metric spaces as follows, 
\begin{itemize}
    \item $(\setS, F)$ is the space of non-singular matrices equipped with the Frobenius distance.
    \item $(\setC, \mathcal{B})$ is the space of positive-definite matrices equipped with the Bures distance.
\end{itemize} 
Both $(\setS, F)$ and $(\setC, \gB)$ have Riemannian structures such that the associated Riemannian gradients can be defined. This section provides simplified main results on the Riemannian geometry, detailed discussions can be found in Appendix \ref{appb1}. 

Given a functional $\gF:\setS \to \mathcal{R}$, the manifold on $\setS$ with the metric tensor $\gG$ by the Frobenius inner product $\langle A, B \rangle_{\gG}=\rm{tr}(A^TB)$ has the Riemannian gradient as,
\begin{equation*}
    \mathrm{grad}_{\gG} \gF(S) = \nabla_S \gF(S).
\end{equation*}
The Riemannian gradient is directly given by the matrix derivative $\nabla_S \gF(S)$. This means optimization algorithms in the space $(\setS, F)$ are straightforward due to its "flat" geometry, akin to $\mathbb{R}^n$. Given this result, we call the ODE in \myeqref{scale_ode} the Euclidean gradient flow of the KL divergence.

\textbf{Remark 3.} The term "Euclidean", while potentially a stretch from its strictest definition, might be slightly abused but is not misleading. The Frobenius inner product extends the concept of Euclidean inner product to the matrix space. Consequently, it inherits characteristics of flat geometry—like those associated with curvatures.

Next, we define a smooth map $ \pi:\setS \to \setC $ as
\begin{equation*}
    \pi(S) = S S^T \in \setC.
\end{equation*}
The differential of this map $\ddpi_S: \gT_S \setS \to \gT_{\pi(S)} \setC $ acts as 
\begin{equation*}
    \ddpi_S(X) = X S^T + S X^T, \quad X \in \gT_S \setS,
\end{equation*}
to map $X$ from the tangent space at $S \in \setS$ to the tangent space at its image $\pi(S) \in \setC$. The differential map is obviously surjective such that the tangent space $\gT_S \setS$ can be decomposed into two subspaces
\begin{equation*}
     \gT_S \setS = \gV_{S}\oplus\gH_{S},
\end{equation*}
where vertical space $\gV_{S}$ is the kernel of the differential map which comprises all elements that are mapped to zeros,
 \begin{equation*}
      \gV_S = \gK\ddpi_S = \{X| X S^T + S X^T=0\},
 \end{equation*}
and the horizontal space $\gH_S$ is the orthogonal complement to $\gV_S$ with respect to $\gG$, given by
\begin{equation*}
    \gH_S = \{X| XS^{-1} \text{ is symmetric}\}.
\end{equation*}
Geometrically, the kernel $\gK\ddpi_S$ represents the directions in which the mapping $\pi(S)$ is locally constant near the point $S\in \setS$.
The decomposition determines that only horizontal vectors $X \in \gH_S$ are mapped to $\gT_{\pi(S)} \setC$.

Suppose that given a metric tensor $\gQ$ on $\setC$, \cite{takatsu2011wasserstein} and \cite{bhatia2019bures} showed that if the map $\pi$ is a Riemannian submersion to satisfy
\begin{equation} \label{submersion}
    \langle \ddpi_S(X), \ddpi_S(Y)\rangle_{\gQ} = \langle X,Y \rangle_{\gG}, \quad \text{for } X,Y \in \gH_S,
\end{equation}
then the distance function induced by this metric tensor $\gQ$ is the Bures distance in \myeqref{w_2_gauss}. This suggests that we can translate the Bures-Wasserstein geometry into a more analytically tractable Euclidean geometry as discussed below.

\begin{proposition}\label{hor} The Euclidean gradient of the KL divergence with respect to the scale matrix $S$ in \myeqref{euc_grad} can be rewritten as
\begin{equation*} 
    \nabla_S \KL(q_\theta||p)=-\eqx \left[ \gradx^2 \log p(\rvx) \right] \cdot S -  S^{-T},
\end{equation*}
and it is horizontal, i.e., $\nabla_S \KL(q_\theta||p) \in \mathcal{H}_{S}$. 
\end{proposition}

\begin{lemma}\label{lemma11} Given two functionals: $\gF: \setS \to \mathcal{R}$ and $\gE: \setC \to \mathcal{R}$ satisfying
$$\gF(S) = \gE(\pi(S)), \quad S \in \setS$$
where the map $\pi$ is the Riemannian submersion satisfying \myeqref{submersion}. If $\mathrm{grad}_{\gG} \gF(S)$ is horizontal, we have
 $$\mathrm{grad}_{\gQ} \gE(\pi(S))= \ddpi_S\big(\mathrm{grad}_{\gG}\gF(S)\big). $$
\end{lemma}

Proofs for \autoref{hor} and \autoref{lemma11} can be found in Appendix \ref{proof_p2} and \ref{proof_l1} respectively. \autoref{hor} shows that the Euclidean gradient of the KL divergence with respect to the scale matrix has no vertical component such that it is exactly mapped to the tangent space $\gT_{\pi(S)} \setC$ under the Riemannian submersion, to reconstruct the Riemannian gradient of the KL divergence with respect to the covariance matrix by \autoref{lemma11}. The Riemannian gradient in $\setC$ is given by
\begin{equation*}
    \begin{split}
    \mathrm{grad}_{\gQ}\KL(q_\theta||p) &=  \ddpi_S \big(\nabla_S \KL(q_\theta||p)\big) \\
    &= -2I - \eqx \left[ \gradx^2 \log p(\rvx) \right]\cdot \Sigma -  \Sigma\cdot \eqx \left[ \gradx^2 \log p(\rvx) \right], \quad \Sigma = SS^T.
\end{split}
\end{equation*}
This Riemannian gradient corresponds to the Hessian form (see \myeqref{hessian_form}) of the ODE system in \myeqref{ode_system}. As a result, the image of an Euclidean gradient flow in $(\setS, F)$ is also a gradient flow of the same functional in $(\setC, \gB)$. Furthermore, as a direct result of the Riemannian submersion, given a curve $\{ \Sigma_t\}_{t \geq0} \in \setC$, for any point $S_0 \in \pi^{-1}(\Sigma_0)$, the curve $\{ S_t\}_{t \geq0} \in \setS$ starting from $S_0$ with $\Sigma_t=S_tS_t^T$ is unique.

The above geometric analysis aligns with the previous result derived from the limiting case of the gradient descent algorithms. For a more in-depth discussion on Riemannian submersion, we direct the reader to \citep{petersen2006riemannian}.
\begin{tcolorbox}[colback=NavyBlue!10,colframe=NavyBlue!15, boxrule=1.5mm, arc=5mm]
The horizontal space $\gH_S = \{X| XS^{-1} \text{ is symmetric}\}$ indicates $\forall X \in \gH_S, XS^{-1}$ is symmetric. In another coordinate system, we can also have $\dd \pi_S (X) = XS^{-1}$ (see Proposition 3.1 by \citealt{takatsu2011wasserstein}), this gives another form of Riemannian gradient in $\setC$ by \autoref{lemma11},
\begin{equation*}
\begin{split}
    \mathrm{grad}_{BW}\KL(q_\theta||p) &=  \ddpi_S \big(\nabla_S \KL(q_\theta||p)\big),\\
    &= -\eqx \left[ \gradx^2 \log p(\rvx) \right] -  \Sigma^{-1}, \quad \Sigma = SS^T,
\end{split}
\end{equation*}
which is called the Bures-Wasserstein gradient studied by \citet{altschuler2021averaging, lambert2022variational, diao2023forward}. Nevertheless, the geometry of the manifold is independent of the choice of coordinate system.
\end{tcolorbox}

\subsection{An Illustrative Example}\label{illus1}
In this part, we provide an illustrative example to see the empirical behaviors of three variational inference algorithms with the Gaussian family: 
\begin{enumerate}
    \item BBVI using the reparameterization gradient, see \autoref{alg}.
    \item BBVI using the path-derivative gradient from \myeqref{path}.
    \item The ODE in Eq.~(\ref{ode_system}) by \cite{lambert2022variational} where we use the forward Euler scheme (gradient descent).
\end{enumerate}
The target distribution is a 2D Gaussian, and the initialization of the variational distribution remains the same across all three algorithms. We employ the Monte Carlo method to evaluate gradients, using the same sample size and learning rate for gradient descent in all cases. In addition, we also simulate the Langevin SDE in \myeqref{langevin_sde}, which is commonly referred to as the Ornstein-Uhlenbeck (OU) process when the target distribution $p(\rvx)$ is Gaussian. In this example, the marginals of the OU process remain Gaussian directly as a result of the Itô integral \citep{wibisono2018sampling}.

\begin{figure}[ht]
\centering
\includegraphics[width=0.8\textwidth]{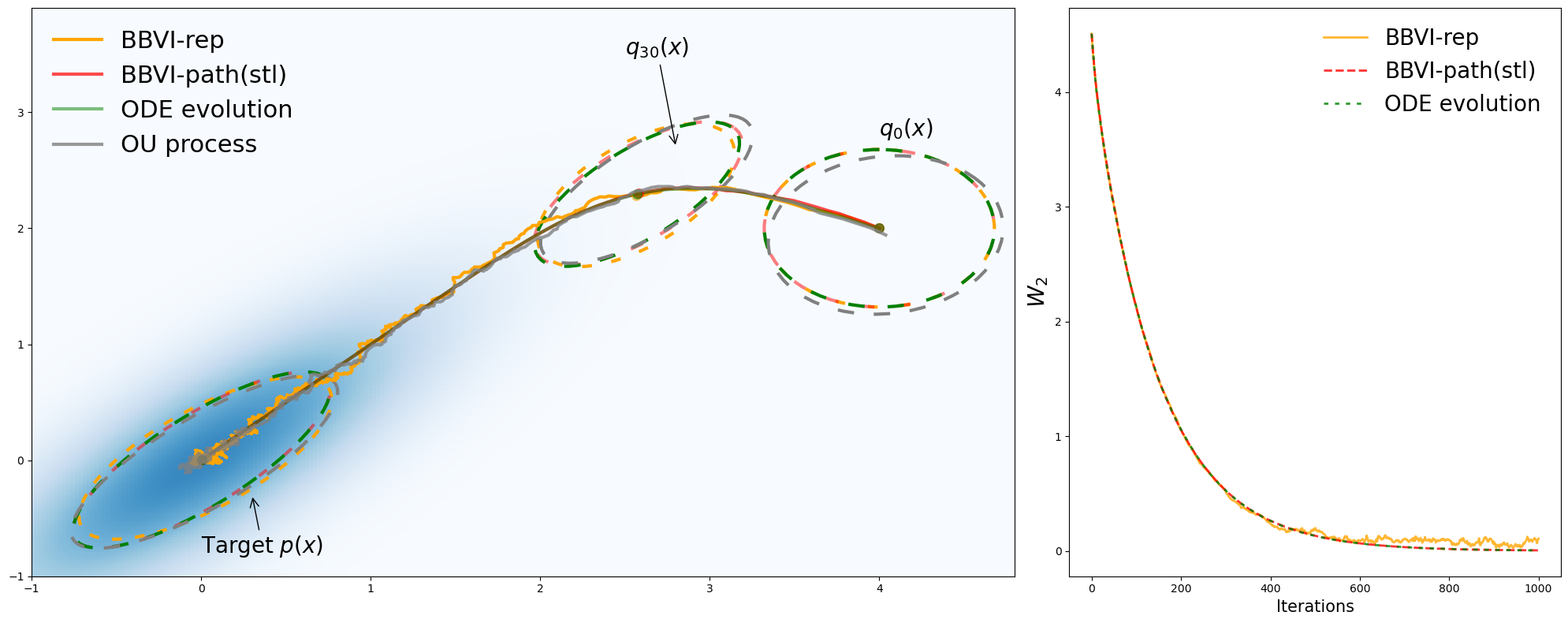}
\caption{Trajectories of means of Gaussian variational distributions with variance ellipsoids at the initial, iteration 30 and the final step. The trajectory and variance ellipsoid for the OU process are evaluated empirically from particles. The right figure is the Wasserstein-2 distance between the variational and the target distributions. It can be seen that "BBVI-path(stl)" and "ODE evolution" have exact convergences without variances.}
\label{illus}
\end{figure}

In \autoref{illus}, we can observe that three algorithms have the same evolutions as well as the OU process (ignoring the errors raised by Monte Carlo sampling and the discretization of gradient flows), especially, we can observe that BBVI with the path-derivative gradient and the ODE evolution both obey "sticking the landing" property (exact convergence without variance in terms of the trajectories and the Wasserstein-2 metrics) \citep{roeder2017sticking}. This is because the vector field (gradient) vanishes if the variational distribution closely approximates $p(\rvx)$. More examples using larger sample sizes for Monte Carlo simulation and non-Gaussian target distributions are included in Appendix \ref{exp1}.

\section{Distillation: An Alternative View Beyond the KL divergence and the Gaussian Family}
We have established the relationship between BBVI and the Bures-Wasserstein gradient flow under the Gaussian variational family and the KL divergence. But how can we generalize it to $f$-divergences and non-Gaussian families? This section provides insights into this question. We first present an equivalent implementation of the path-derivative gradient using an iterative distillation procedure. We then extend this distillation to general $f$-divergences, leading to a novel gradient estimator. We demonstrate that this new estimator is statistically unbiased.

\subsection{Distillation: From Sample Space to Parameter Space}
We may have noticed that the path-derivative gradient in Eq.~(\ref{stl}) generates the Wasserstein gradient of the KL divergence, 
\begin{equation*}
    \dtheta \KL(q_\theta||p) = \ee_{\rvz\sim \xi} \big[\wgrad \fkl(q_\theta)(\rvx_\theta) \circ \dtheta \rvx_\theta\big],
\end{equation*}
where $ -\wgrad \fkl(q_\theta)(\rvx) =  \gradx \big(\log p(\rvx) /q(\rvx;\theta)  \big)$ represents the vector field of the probability flow ODE of the KL divergence in \myeqref{prob_ode}. In this section, we will show that the path-derivative of the KL divergence can be equally implemented via an iterative distillation procedure, also known as the amortization trick \citep{wang2016learning, yi2023monoflow}.

First, suppose that at step $k$, $q^\tau_{k}$ represents the marginal distribution of the probability flow ODE in \myeqref{prob_ode} with a parametric density function $q(\rvx;\theta)$ and $v_k$ represents the vector field of the ODE, written as 
\begin{equation*}
    v_k(\rvx)= \gradx \big(\log p(\rvx) - \log q(\rvx;\theta)  \big),
\end{equation*}
and particles are reparameterized by $\rvx_{\theta} = g(\rvz;\theta) \sim q^\tau_{k}$, $\rvz \sim \pz$. Next, we consider moving particles along the vector field with the step size $\tau$, i.e., one-step forward Euler scheme, 
\begin{equation*}
    \rvx' = \rvx_{\theta} + \tau v_k(\rvx_{\theta}).
\end{equation*}
In the space of probability distributions, this iteration corresponds to 
\begin{equation*}
    q^\tau_{k+\frac{1}{2}} = \big[\text{id} - \tau \wgrad  \fkl(q_\theta)\big]_{\#} q^\tau_k,
\end{equation*}
where $\#$ is the pushforward operator. The above forward proceeding operation is depicted in the following diagram, 
\[\begin{tikzcd}[column sep={7em}, row sep=small, arrows={decorate, decoration={snake,segment length=2.5mm, amplitude=0.5mm}},]
\textbf{Samples: \quad  } &[-15em]&\rvx_{\theta} \arrow[r, "v_k", decorate=false] \arrow[d, dash, decorate=true]   & \rvx' \arrow[d, dash,decorate=true]\\
\textbf{Marginals:   \quad   }&[-3em]& q^\tau_k\arrow[r,"-\wgrad  \fkl(q_\theta)", decorate=false]& q^\tau_{k+\frac{1}{2}}
\end{tikzcd}\]
\begin{wrapfigure}[19]{r}{0.48\textwidth}
    \includegraphics[width=0.45\textwidth]{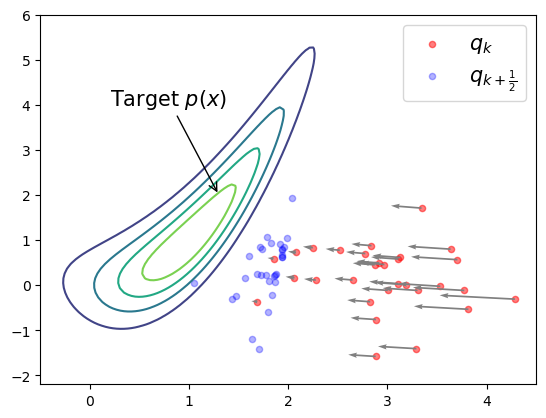}
    \caption{The target $p(\rvx)$ is a banana distribution following the Rosenbrock density \citep{rosenbrock1960automatic}. If given $q^\tau_k$ as a Gaussian distribution, applying the Wasserstein gradient no longer guarantees particles from $q^\tau_{k+\frac{1}{2}}$ maintains Gaussian, see the blue particles.}
     \label{demo}
 \end{wrapfigure}

Notice that $q^\tau_{k+\frac{1}{2}}$ has no closed-form density generally, and it is only represented by some particles $\rvx'$, see \autoref{demo}. 

The second step is to find a distribution $q^\tau_{k+1} = q(\rvx;\theta_\text{new})$ that numerically approximates $q^\tau_{k+\frac{1}{2}}$. Since the sample space is $\mathbb{R}^n$, a naive approach is to minimize the squared Euclidean distance between $\rvx'$ and $\rvx_\theta$ via a single-step gradient descent,
\begin{equation*}
\begin{split}
    &\theta_{\text{new}} \leftarrow \theta_\text{old}-  \dtheta l(\theta)|_{\theta=\theta_\text{old}}, \\
    & \text{where }l(\theta) = \frac{1}{2}\mathbb{E}_{\rvz \sim \pz} \big[\Vert \rvx_{\theta}-\rvx_s'\Vert^2_2\big],
    \end{split}
\end{equation*}
where $\rvx'_s$ means that the stop gradient operator is applied to $\rvx'$ to discard the computational graph on $\theta$, i.e., $\rvx'_s$ becomes a fixed constant. Minimizing the loss function $l(\theta)$ encourages $g(\rvz;\theta)$ to draw particles as similar to $\rvx'$ as possible. For example, in \autoref{demo}, $g(\rvz;\theta)$ is encouraged to learn to draw the blue particles at $q^\tau_{k+\frac{1}{2}}$ which are closer to the target distribution.
Applying $\dtheta$ to $l(\theta)$, we obtain
\begin{equation}
\label{kl_distill}
    \dtheta l(\theta) = \mathbb{E}_{\rvz \sim \pz} \big[ (\rvx_\theta - \rvx'_s) \circ \dtheta \rvx_\theta \big] 
    = \tau \ee_{\rvz\sim \xi} \big[\wgrad\fkl(q_\theta)(\rvx_\theta) \circ \dtheta \rvx_\theta\big]. 
\end{equation}
This shows that $\dtheta l(\theta)$ is equal to the path-derivative gradient of the KL divergence up to the step size $\tau$, which means doing distillation is identical to performing BBVI with the path-derivative gradient. The difference here is that the stop gradient operator is applied to the updated particles $\rvx'$ instead of the variational parameter $\theta$. We summarize the distillation procedure as per iteration $k$:
\begin{enumerate}
    \item sample particles $\rvx_\theta$ from $q^\tau_k=q(\rvx;\theta)$, and move particles via $\rvx' = \rvx_{\theta} + \tau v_k(\rvx_{\theta})$.
    \item apply the stop gradient operator to particles $\rvx'$ and evaluate the loss $l(\theta)$.
    \item backpropagate the loss $l(\theta)$ and update $\theta_{\text{new}} \leftarrow \theta_\text{old}-  \dtheta l(\theta)|_{\theta=\theta_\text{old}}$
\end{enumerate}
The advantage of the distillation procedure is that it only relies on vector fields given by Wasserstein gradients and does not require explicit forms of the divergence and the variational family.

\subsection{Distilling the Probability Flow ODE of $f$-Divergence}
 In this section, we extend distillation to the probability flow ODE of $f$-divergence.  The $f$-divergence is defined as
\begin{equation*}
 \mathcal{F}_{f}(q)=\mathcal{D}_{f}(p||q) =  \int  f\left(\frac{p(\rvx)}{q(\rvx)} \right)q(\rvx) \dd \rvx,
\end{equation*}
where $f$ is a convex function with $f(1)=0$. 

In order to apply the distillation procedure, we need to obtain the vector field for the probability flow ODE of $f$-divergence. Recall that in \myeqref{ode_vec}, the minus vector field is the Wasserstein gradient, which is given by the Euclidean gradient of the first variation. \autoref{first_var} offers an explicit expression of the first variation of $f$-divergences.

\begin{lemma}\label{first_var}
    The first variation of $\mathcal{F}_{f}(q)$ is given by 
\begin{equation*}
    \frac{\delta \mathcal{F}_{f}(q)}{\delta q} = f(r) - r f'(r), \quad r = \frac{p}{q}.
\end{equation*}
\end{lemma} 
The proof of \autoref{first_var} is provided in Appendix \ref{le1}  or alternatively, see Theorem 3.2 by \cite{yi2023monoflow}.
By \autoref{first_var}, the Wasserstein gradient of $f$-divergence is given by
\begin{equation*}
    \nabla_{W_2}\mathcal{F}_{f}(q)  = \gradx \big[ f(r) - r f'(r) \big],
\end{equation*}
where the associated probability flow ODE is characterized by 
\begin{equation}
    \dd \rvx_t = \gradx \big[ r_t(\rvx) f'(r_t(\rvx_t)) - f(r_t(\rvx_t)) \big] \dd t, \quad r_t = \frac{p}{q_t}.\label{f_pro_ode}
\end{equation}
Following the previous distillation procedure, if we move particles $\rvx_\theta$ along the vector field in \myeqref{f_pro_ode} to $\rvx'$ and evaluate the quadratic Euclidean distance $l(\theta)$ between them, we can obtain a novel gradient estimator by replacing $\wgrad\fkl(q)$ with $\nabla_{W_2}\mathcal{F}_{f}(q)$ in Eq.~(\ref{kl_distill})  such that
\begin{equation}\label{wass_armo}
\begin{split}
    \dtheta l(\theta) \propto \eez \big[\wgrad\gF_f(q_\theta)(\rvx_\theta) \circ \dtheta \rvx_\theta\big].
\end{split}
\end{equation}
Similar to the path-derivative of the KL divergence \citep{roeder2017sticking}, \myeqref{wass_armo} can be realized by applying the stop gradient operator to the parameter $\theta$ directly. By reorganizing \myeqref{wass_armo}, we obtain
\begin{equation}\label{res}
         \dtheta l(\theta) \propto -\ee_{\rvz\sim \xi} \big[ \gradx h(r(\rvx_\theta;\theta))  \circ \dtheta \rvx_\theta\big]
     = - \eez \left[ \dtheta h\big(\ratios\big)\right]
\end{equation}
where $h(r) = rf'(r) - f(r)$, and $r(\rvx;\theta_s) = p(\rvx)/q(\rvx;\theta_s)$. It is obvious that \myeqref{res} follows the straightforward result of the chain rule, or see \myeqref{62} in Appendix \ref{path app} for more discussions on the stop gradient operator. 


\subsubsection{Statistical Unbiasedness}
The distillation procedure is simply based on heuristics. In this section, we show that the previously obtained estimator in \myeqref{res} is an unbiased gradient estimator of $f$-divergence $\mathcal{D}_{f}(p||q_\theta)$ with respect to the parameter $\theta$. We call this estimator the path-derivative gradient of $f$-divergence.

\begin{proposition}\label{p3}
Given the reparameterization, 
$$\rvx_\theta = g(\rvz;\theta) \sim q(\rvx;\theta), \rvz \sim \pz,$$  
the path-derivative gradient estimator of $f$-divergences is given by,
\begin{equation*}
    \dtheta \mathcal{D}_{f}(p||q_\theta) =  - \eez \left[ \dtheta h\big( r(\rvx_\theta; \theta_{s})\big) \right], \quad r(\rvx;\theta_s) = \frac{p(\rvx)}{q(\rvx;\theta_s)}
\end{equation*}
where $h\colon \mathcal{R}^+ \to \mathcal{R}$ satisfies $h(r) = rf'(r) - f(r)$ and $\theta_s$ means the stop gradient operator is applied. 
\end{proposition} 
The proof of \autoref{p3} can be found in Appendix \ref{proof_p3}. It indicates that distilling the probability flow ODE of $f$-divergence is exactly equivalent to variational inference problems that use gradient descent to update parameters. Recall from Eq.~(\ref{rep_grad}) where we have the reparameterization gradient of the KL divergence, similarly we can use this trick to obtain the reparameterization gradient of $f$-divergence (by using the law of unconscious statisticians and  the interchange between differentiation and integration, see Appendix \ref{appb}) such that we have
\begin{equation*}
    \dtheta \mathcal{D}_{f}(p||q_\theta)=\color{red}{\underbrace{ \color{black}\mathbb{E}_{\rvz \sim \pz} \left[\dtheta f\big(\ratio \big)\right]}_\text{reparameterization gradient}}\color{black}
    =\color{blue}\underbrace{\color{black}- \mathbb{E}_{\rvz \sim \pz}\left[ \dtheta h\big( r(\rvx_\theta; \theta_{s})\big) \right]}_\text{path-derivative gradient} \color{black}. \label{gradf}
\end{equation*}
The difference between these two estimators is that they are evaluated on entirely different Monte Carlo objectives. The reparameterization gradient requires a convex function $f$ of the density ratio $r(\rvx;\theta) = p(\rvx)/q(\rvx;\theta)$ that is differentiable with respect to both sample $\rvx$ and parameter $\theta$. The path-derivative gradient requires a non-decreasing function $h$ of density ratio $r(\rvx;\theta_s)$ which is a function only differentiable with $\rvx$. 

The path-derivative gradient also defines a surrogate loss function $L(\theta)$ which allows us to perform BBVI via
\begin{equation}\label{surro}
    \min_\theta L(\theta)=- \eez \left[ h\big(\ratios\big)\right], \quad \rvx_\theta = g(\rvz;\theta) \sim q(\rvx;\theta), \rvz \sim \pz
\end{equation}

\textbf{Remark 4.} The convexity of $f$ indicates that $h$ is a non-decreasing function, due to that $h(r)=rf'(r) - f(r) \Longrightarrow h'(r) = rf''(r) \geq 0$. If $f$ is strictly convex which implies $f''(r)>0$, the associated $h$ is strictly increasing. In generative adversarial nets (GANs), \cite{yi2023monoflow} showed that the generator loss of divergence GANs follows
\begin{equation} \label{gans}
    - \eez \left[ h\big( \hat{r}(\rvx_\theta) )\right], \quad \rvx_\theta=g(\rvz;\theta)\sim p_\text{generator},
\end{equation}
where $h$ can be an arbitrary increasing function and $\hat{r}(\rvx)$ is the density ratio estimator of $\pd(\rvx)/p_\text{generator}(\rvx)$. The difference between \myeqref{surro} and \myeqref{gans} is that $\hat{r}(\rvx)$ is obtained by two-sample density ratio estimation \citep{sugiyama2008direct, moustakides2019training}, and $r(\rvx;\theta_s)$ is obtained by applying the stop gradient operator to $\theta$ in the ground truth density ratio. It can be seen that both $r(\rvx;\theta_s)$ and $\hat{r}(\rvx)$ are functions only with the variable $\rvx$.

\subsubsection{Special Cases}
The path-derivative gradient of $f$-divergence generalizes several gradient estimators.
\begin{itemize}
    \item \textbf{(Reverse) KL divergence}: $f(r) = -\log r \Longrightarrow h(r) = \log r -1 $, 
     \begin{equation}
     - \mathbb{E}_{\rvz \sim \pz}\left[ \dtheta h\big( r(\rvx_\theta; \theta_{s})\big) \right] = \mathbb{E}_{\rvz \sim \pz} \Big[ \dtheta \log\big[q(\rvx_\theta;\theta_s)/{p}(\rvx_{\theta})\big]\Big],
 \end{equation}
  we obtain the "sticking the landing" estimator \citep{roeder2017sticking}.
  \item \textbf{Forward KL divergence}: $f(r) = r\log r \Longrightarrow h(r) = r ,$
   \begin{equation}
     - \mathbb{E}_{\rvz \sim \pz}\left[ \dtheta h\big( r(\rvx_\theta; \theta_{s})\big) \right] = -\mathbb{E}_{\rvz \sim \pz} \Big[ \dtheta \big[p(\rvx_{\theta}) /q(\rvx_\theta;\theta_s)\big]\Big]
 \end{equation}
 this recovers the gradient estimator by \cite{vaitl2022gradients}.
 \item \textbf{$\alpha$-divergence ($\alpha\neq0$)}: $f(r) = \frac{r^\alpha-\alpha r - (1-\alpha)}{\alpha(\alpha-1)} \Longrightarrow h(r) = \frac{r^\alpha -1}{ \alpha}$, 
 \begin{equation}
     - \mathbb{E}_{\rvz \sim \pz}\left[ \dtheta h\big( r(\rvx_\theta; \theta_{s})\big) \right] 
     = -\mathbb{E}_{\rvz \sim \pz} \left[ \frac{1}{\alpha}\dtheta\left(\frac{p(\rvx_{\theta})} {q(\rvx_\theta;\theta_s)}\right)^{\alpha}\right],
\end{equation}
this recovers the gradient estimator by \cite{geffner2020difficulty}.
\end{itemize}
For an unnormalized density $p(\rvx)$ where $p_{\text{true}}(\rvx) = p (\rvx) / C$, we have 
$$
\left(\frac{p(\rvx)} {q(\rvx;\theta_s)}\right)^{\alpha} \propto \left(\frac{p_{\text{true}}(\rvx)} {q(\rvx;\theta_s)}\right)^{\alpha},
$$
this shows that the normalizing constant only affects the scale of the gradient if the given divergence belongs to $\alpha$-divergence family.

\begin{figure*}[t!]
\vspace{-0.0cm}
     \centering
     \begin{subfigure}{0.18\textwidth}
         \centering
         \includegraphics[width=\textwidth]{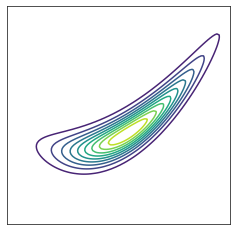}
          \caption{Target $p(\rvx)$}
         \vspace{-0mm}
     \end{subfigure}
     \begin{subfigure}{0.18\textwidth}
         \centering
         \includegraphics[width=\textwidth]{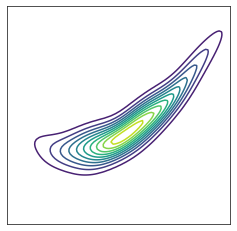}
          \caption{Reverse KL}
         \vspace{-0mm}
     \end{subfigure}
     \begin{subfigure}{0.18\textwidth}
         \centering
         \includegraphics[width=\textwidth]{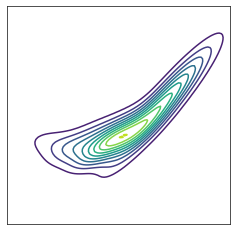}
         \caption{Forward KL}
         \vspace{-0mm}
     \end{subfigure}
      \begin{subfigure}{0.18\textwidth}
         \centering
         \includegraphics[width=\textwidth]{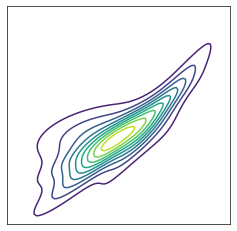}
         \caption{$\chi^2$}
         \vspace{-0mm}
     \end{subfigure}
      \begin{subfigure}{0.18\textwidth}
         \centering
         \includegraphics[width=\textwidth]{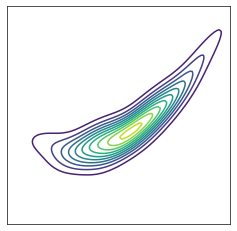}
         \caption{Hellinger}
         \vspace{-0mm}
     \end{subfigure}
     \caption{Contour plots of the target distribution, and the approximated Gaussian mixture models by minimizing various $f$-divergences (see \autoref{expilcit form} in Appendix \ref{proof_p3}) using the path-derivative gradient.}
     \label{funcs}
\end{figure*}

In \autoref{funcs}, we implement the path-derivative gradient for the Gaussian mixture variational family to approximate an unnormalized density function given by the Rosenbrock function \citep{rosenbrock1960automatic} under different $f$-divergences. More empirical evaluations can be found in Appendix \ref{final_exp},  such as the comparison between the reparameterization and the path-derivative gradients, and Bayesian logistic regression on the UCI dataset \citep{asuncion2007uci}. 

 \section{Related Works}
\textbf{Bures-Wasserstein geometry}. \cite{lambert2022variational} first studied the variational inference problem under the Bures-Wasserstein geometry such that the mean and covariance evolution can be derived, subsequent work by \cite{diao2023forward} investigated the forward-backward scheme to address the non-smoothness of the KL divergence. Together with another work by \cite{altschuler2021averaging} which studied the Bures-Wasserstein space for barycenter problems, all of these works consider parameterizing Gaussians with covariances such that the optimization algorithms are derived in a non-Euclidean space. The connection between the Euclidean space of non-singular matrices and the Bures space of positive-definite matrices was first established by \cite{takatsu2011wasserstein,  modin2016geometry, bhatia2019bures}. Based on that, we showed that the gradient of the KL divergence w.r.t. the scale matrix of Gaussians is horizontal under the Riemannian submersion. This bypasses the difficulty in dealing with the non-Euclidean geometry and also demonstrates that conventional VI methods naturally involve the Wasserstein geometry. 

 \textbf{Variational inference and path-derivative gradients}. The standard variational inference methods consider the problem of minimizing the reverse KL divergence \citep{jordan1998variational, kingma2013auto,hoffman2013stochastic, rezende2014stochastic, blei2017variational}. Minimizing the reverse KL divergence often results in mode-seeking tendencies and underestimates the uncertainties in the target distribution. To address this issue, other classes of $f$-divergence have also been studied, e.g., the forward KL divergence \citep{minka2013expectation, naesseth2020markovian, jerfel2021variational, vaitl2022gradients}, the $\alpha$-divergence \citep{hernandez2016black, li2016renyi, geffner2020difficulty}. The majority of these VI methods are based on deriving specific evidence bounds to obtain a gradient estimator that allows for optimization whereas our gradient estimator is divergence-agnostic. We also noticed that our gradient estimator of $f$-divergence generalizes several works designing specific estimators using the stop gradient operator, e.g., the reverse KL divergence \citep{roeder2017sticking}, the forward KL divergence \citep{vaitl2022gradients}, $\alpha$-divergences \citep{geffner2020difficulty}. To the best of our knowledge, the path-derivative gradient we introduced is the most unified form. The path-derivative gradient estimator relies on a stop-gradient operator, this operator also arises in importance-weighted variational objectives \citep{tucker2018doubly, finke2019importance}, doubly-reparameterized gradient \citep{bauer2021generalized}, normalizing flow models \citep{agrawal2020advances, vaitl2022path} and a low variance VI approach \citep{richter2020vargrad}. 

\section{Discussion}
In this paper, we bridge the gap between variational inference and Wasserstein gradient flows. Under certain conditions (Gaussian and the KL), we showed that the Bures-Wasserstein gradient flow can be obtained via the Euclidean gradient flow where its forward scheme is exactly the black-box variational inference algorithm. This equivalence is also a result of the Riemannian submersion which maps the Euclidean gradient to the Riemannian gradient in another space. We further showed that beyond the Gaussian family and the KL divergence, by distilling the Wasserstein gradient flows, we also obtained a new gradient estimator that is statistically unbiased. However, while the Gaussian variational family's geometry is more straightforward, analyzing the geometry for a general parameter space remains to be challenging. Additionally, the variance analysis of the path-derivative gradient is still an unresolved matter.

\bibliography{ref}
\bibliographystyle{plainnat}
\newpage
\appendix
\addcontentsline{toc}{section}{Appendix} 
\part{Appendix} 
\parttoc 
\newpage
\section{Equivalent Formulations of the ODE System} \label{equal_ode}
Given $q_t = \mathcal{N}(\mu_t, \Sigma_t)$, the ODE system following the Bures-Wasserstein gradient flow \citep{lambert2022variational} is given by
\begin{equation*}
\begin{split}
    &\frac{\dd \mu_t}{\dd t} =  \mathbb{E}_{\rvx \sim q_t}\left[ \gradx \log \frac{p(\rvx)}{ q_t(\rvx)}\right],\\
    &\frac{\dd \Sigma_t}{\dd t} =   \mathbb{E}_{\rvx \sim q_t}\left[\left(\gradx \log \frac{p(\rvx)}{ q_t(\rvx)} \right)^T(\rvx - \mu_t) \right] +\mathbb{E}_{q_t}\left[ (\rvx-\mu_t)^T \gradx \log \frac{p(\rvx)}{ q_t(\rvx)} \right].
\end{split}
\end{equation*}

If the target distribution is an energy distribution $p(\rvx)= \exp\big[-V(\rvx) \big]$,
the mean evolution can be written as 
\begin{equation}
    \frac{\dd \mu_t}{\dd t} =  \mathbb{E}_{\rvx \sim q_t}\big[ \gradx \log p(\rvx) \big] = -\mathbb{E}_{\rvx \sim q_t}\big[ \gradx V(\rvx) \big]\label{eq45}
\end{equation}
by the fact $\mathbb{E}_{\rvx \sim q_t}\big[ \gradx \log q_t(\rvx) \big]=0$ since $q_t$ is a Gaussian.\\

Using $\rvx - \mu_t= -\gradx \log q_t(\rvx) \cdot \Sigma_t$, the covariance evolution can be written as,
\begin{equation*}
\begin{split}
        \frac{\dd \Sigma_t}{\dd t} &=   \mathbb{E}_{\rvx \sim q_t}\left[\left(\gradx \log \frac{p(\rvx)}{ q_t(\rvx)} \right)^T(\rvx - \mu_t) \right] +\mathbb{E}_{q_t}\left[ (\rvx-\mu_t)^T \gradx \log \frac{p(\rvx)}{ q_t(\rvx)} \right]\\
        & =  -\mathbb{E}_{\rvx \sim q_t}\left[\left(\gradx \log \frac{p(\rvx)}{ q_t(\rvx)} \right)^T\gradx \log q_t(\rvx) \right]\cdot \Sigma_t -\Sigma_t^T \cdot \mathbb{E}_{q_t}\left[ \gradx \log q_t(\rvx)^T \gradx \log \frac{p(\rvx)}{ q_t(\rvx)} \right].
\end{split}
\end{equation*}
Using  integral by part, we have
\begin{equation*}
    \mathbb{E}_{\rvx \sim q_t}\left[\left(\gradx \log  q_t(\rvx) \right)^T\gradx \log q_t(\rvx) \right]\cdot \Sigma_t = 0 - \mathbb{E}_{\rvx \sim q_t}\left[\gradx^2 \log q_t(\rvx) \right]\cdot \Sigma_t = I.
\end{equation*}
Hence the covariance evaluation can also be written as 
\begin{equation}
    \frac{\dd \Sigma_t}{\dd t} = 2I -  \mathbb{E}_{\rvx \sim q_t}\left[\left(\gradx V(\rvx) \right)^T(\rvx - \mu_t) + (\rvx-\mu_t)^T \gradx V(\rvx) \right],\label{eq48}
\end{equation}
\begin{tcolorbox}[colback=NavyBlue!10,colframe=NavyBlue!15, boxrule=1.5mm, arc=5mm]
Combing Eq.~(\ref{eq45}) and Eq.~(\ref{eq48}), we have 
\begin{equation*}
\begin{split}
     \frac{\dd \mu_t}{\dd t} &=  -\mathbb{E}_{\rvx \sim q_t}\big[ \gradx V(\rvx) \big]\\
      \frac{\dd \Sigma_t}{\dd t} &= 2I -  \mathbb{E}_{\rvx \sim q_t}\left[\left(\gradx V(\rvx) \right)^T(\rvx - \mu_t) + (\rvx-\mu_t)^T \gradx V(\rvx) \right],
\end{split}
\end{equation*}
 this is known as the Sarkka equation in Bayesian filtering \citep{sarkka2007unscented}. 
 \end{tcolorbox}
 \begin{tcolorbox}[colback=NavyBlue!10,colframe=NavyBlue!15, boxrule=1.5mm, arc=5mm]
 Using integral by part again, we obtain an equivalent Hessian form,
\begin{equation}
\begin{split}
     \frac{\dd \mu_t}{\dd t} &=  -\mathbb{E}_{\rvx \sim q_t}\big[ \gradx V(\rvx) \big]\\
      \frac{\dd \Sigma_t}{\dd t} &= 2I -  \Sigma_t\cdot\eqt\left[\gradx^2 V(\rvx)\right]  - \eqt\left[\gradx^2 V(\rvx) \right]\cdot \Sigma_t \label{hessian_form}
\end{split}
\end{equation}
\end{tcolorbox}
\newpage

\section{The Path-Derivative Gradient of the KL Divergence} \label{appb}
In order to derive the path-derivative gradient \citep{roeder2017sticking}, we first present two preliminary results: the law of the unconscious statistician (LOTUS) and the interchange between differentiation and integration.
\subsection{Law of the Unconscious Statistician (LOTUS)}

LOTUS offers a straightforward method for computing the expectation under the change of variables. If there exists such a transformation (reparameterization), $$\rvx_\theta = g(\rvz;\theta) \sim q(\rvx;\theta), \rvz \sim \pz.$$ 
By LOTUS, if given a function $\gamma \colon \mathbb{R}^n \to \mathbb{R}^m$, we have the following equality,
\begin{equation*}
    \mathbb{E}_{\rvx \sim q_\theta} [\gamma(\rvx)] = \mathbb{E}_{\rvz \sim \pz} [\gamma(\rvx_\theta)] = \mathbb{E}_{\rvz \sim \pz} [\gamma(g(\rvz;\theta))].
\end{equation*}

\subsection{Interchange Between Differentiation and Integration}
Given $\rvx \in \mathcal{X}$ , $\theta \in \Theta$ and a function $\psi: \mathcal{X} \times \Theta \to \mathcal{R} $.
The availability of the interchange between differentiation and integration
\begin{equation*}
    \frac{\partial }{\partial \theta} \int \psi(\rvx;\theta) \dd \rvx =  \int \frac{\partial }{\partial \theta} \psi(\rvx;\theta) \dd \rvx,
\end{equation*}
holds if the following conditions are true,
\begin{itemize}
    \item $\psi(\rvx;\theta)$ is differentiable with respect to $\theta$ , for almost all $\rvx \in \mathcal{X}$.
    \item $\psi(\rvx;\theta)$ is Lebesgue-integrable with respect to $\rvx$ , for all $\theta \in \Theta$.
    \item There exists a Lebesgue-integrable function $g:\mathcal{X}\to \mathcal{R} $ such that all $\theta \in \Theta$ and almost all $\rvx \in \mathcal{X}$, the following inequality holds,
    $$
        \left\Vert \frac{\partial}{\partial\theta}\psi(\rvx;\theta)  \right\Vert_1 \leq g(\rvx).
    $$
\end{itemize}
These conditions are generally true in machine learning applications, we refer to \citep{mohamed2019monte} for more details.

\subsection{The Path-Derivative Gradient} \label{path app}
For a multivariate function $b(\rvx_\theta; \rvy_\theta):\mathcal{X} \times \mathcal{Y} \to \mathcal{R}$, its derivative with respect to $\theta$ is given by the chain rule,
$$\dtheta b(\rvx_\theta;\rvy_\theta) = \nabla_\rvx b(\rvx_\theta; \rvy_\theta) \circ \nabla_\theta \rvx_\theta + \nabla_\rvy b(\rvx_\theta; \rvy_\theta) \circ \nabla_\theta \rvy_\theta.$$

Therefore, under the reparameterization $\rvx_\theta = g(\rvz;\theta), \rvz \sim \xi $, we have
\begin{equation}
\begin{split}
        \dtheta b(\rvx_\theta;\theta) &= \gradx b(\rvx_\theta;\theta) \circ\dtheta \rvx_\theta + \dtheta b(\rvx;\theta)\vert_{\rvx=\rvx_\theta} \circ\dtheta \theta\\
        & = \gradx b(\rvx_\theta;\theta)\circ \dtheta \rvx_\theta + \dtheta b(\rvx;\theta)\vert_{\rvx=\rvx_\theta}\label{61}.
\end{split}
\end{equation}
In the second term $\dtheta b(\rvx;\theta)\vert_{\rvx=\rvx_\theta}$, the differentiation operator $\dtheta$ works only with respect to the variational parameter $\theta$. \\
\begin{tcolorbox}[colback=NavyBlue!10,colframe=NavyBlue!15, boxrule=1.5mm, arc=5mm]
If we apply the stop gradient operator to $\theta$, this means the differentiation through the variational parameter $\theta$ is discarded  such that we have 
\begin{equation}
    \dtheta b(\rvx_\theta;\theta_s) = \gradx b(\rvx_\theta;\theta) \circ \dtheta \rvx_\theta \label{62}
\end{equation}
In terms of the algorithmic implementation, the stop gradient operator stops backpropagating losses into $\theta$, this means $\theta_s$ becomes a non-trainable variable just like a constant.
\end{tcolorbox}

Based on the above results, we now derive \textbf{the path-derivative gradient of the KL divergence} \citep{roeder2017sticking}.

Let $b(\rvx;\theta) = \log \big[q(\rvx;\theta) /p(\rvx)\big]$ such that
\begin{equation}
\begin{split}
    \dtheta \KL(q_\theta||p) =  \mathbb{E}_{\rvz \sim \pz} \left[\dtheta b(\rvx_\theta;\theta) \right]
    &= \mathbb{E}_{\rvz \sim \pz} \big[ \gradx b(\rvx_\theta;\theta) \circ \dtheta \rvx_\theta + \dtheta b(\rvx;\theta)\vert_{\rvx=\rvx_\theta} \big] \\
    & = \eez \big[ \gradx b(\rvx_\theta;\theta) \circ \dtheta \rvx_\theta \big] +  \eqx \big[\dtheta \log q(\rvx;\theta) \big], \text{by LOTUS}.\\
    & = \eez \big[ \gradx  b(\rvx_\theta;\theta) \circ \dtheta \rvx_\theta \big]+0\\
    & = \eez \big[ \dtheta  b(\rvx_\theta;\theta_s) \big] = \eez \left[ \dtheta  \log \frac{q(\rvx_\theta;\theta_s)} {p(\rvx_\theta)} \right]
\end{split}
\end{equation}
\textbf{Remark:} the score function $\dtheta \log q(\rvx;\theta)$ has a zero mean by
\begin{equation*}
    \eqx \big[\dtheta \log q(\rvx;\theta) \big] = \int   \dtheta  q(\rvx;\theta) \dd \rvx= \dtheta \int  q(\rvx;\theta) \dd \rvx =  \dtheta (1) = 0.
\end{equation*}
 The score function here refers to the derivative of the log density with respect to the parameter $\theta$, which is an enduring terminology in statistical inference. Alternatively, in the context of score matching methods \citep{hyvarinen2005estimation, vincent2011connection}, $\nabla_\rvx \log q(\rvx;\theta)$ is also referred to as the score function. The latter one is commonly used for score-based diffusion models \citep{song2020score}.

\subsection{Proof of \autoref{rule}} \label{proof_p1}
If $q(\rvx;\theta)=\mathcal{N}(\mu, \Sigma)$ with $\Sigma=SS^T$ is a Gaussian distribution with the parameter $\theta = (\mu, S)$ and the reparameterization is given by $\rvx_\theta = \mu + \rvz S^T, \rvz \sim \mathcal{N}(0, I)$.

Specifications for dimensions:
\begin{equation*}
\begin{split}
     &\rvx, \mu, \rvz:1 \times n \text{\ \  row vectors}, \\
    &S :n\times n \text{\ \  matrix.}
\end{split}
\end{equation*}
The path-derivative gradient estimator (sticking the landing) \citep{roeder2017sticking} of the KL divergence is 
\begin{equation*}
   \dtheta \KL(q_\theta||p) =  \mathbb{E}_{\rvz \sim \pz} \left[ \gradx \log \frac{q(\rvx_\theta;\theta_s)}{p(\rvx_\theta)} \circ \dtheta \rvx_\theta \right]. \label{eq41}
\end{equation*}

\textbf{The gradient w.r.t. $\mu$,}\\

The Jacobian $\nabla_\mu \rvx_\theta$ is
\begin{equation*}
    \nabla_\mu \rvx_\theta = I,
\end{equation*}
this is the standard Jacobian of vector-to-vector mappings.
By the chain rule, the gradient w.r.t. $\mu$ is
\begin{equation*}
\begin{split}
    \nabla_\mu \KL(q_\theta||p)  = \mathbb{E}_{\rvz \sim \pz}\left[\nabla_\mu \log \frac{q(\rvx_\theta;\theta_s)}{p(\rvx_\theta)}\right]& =  \mathbb{E}_{\rvz \sim \pz} \left[ \gradx \log \frac{q(\rvx_\theta;\theta_s)}{p(\rvx_\theta)}\cdot  I \right] \\
    & =\mathbb{E}_{\rvz \sim \pz} \left[ \gradx \log \frac{q(\rvx_\theta;\theta)}{p(\rvx_\theta)}\right]\\
    & =- \mathbb{E}_{\rvx \sim q_\theta} \left[ \gradx \log \frac{p(\rvx)}{ q(\rvx;\theta)} \right], \text{by LOTUS}.
\end{split}
\end{equation*}
In the above equation, we can replace $\theta_s$ with $\theta$ since the operator $\gradx$ is irrelevant to $\theta$. 
\\\\
\textbf{The gradient w.r.t. $S$,}

$\nabla_S \rvx_\theta$ is the Jacobian of matrix-to-vector mappings, its dimension is $(1\times n) \times (n\times n)$, its element can be written
\begin{equation*}
\begin{split}
       & \frac{\partial \rvx_i}{\partial S_{i, k}} = \rvz_k, \\
& \frac{\partial \rvx_i}{\partial S_{j, k}} = 0, \quad \text{if } i\neq j\\
\end{split}
\end{equation*}

Similarly, the element-wise derivative w.r.t. the scale matrix is
\begin{equation*}
\begin{split}
    \nabla_S \KL(q_\theta||p)_{i, k} = \eez \left[ \nabla_S \log \frac{q(\rvx_\theta;\theta_s)}{p(\rvx_\theta)}\right]_{i, k} &=\eez\left[ \nabla_{\rvx_i} \log \frac{q(\rvx_\theta;\theta_s)}{p(\rvx_\theta)} \cdot \frac{\partial \rvx_i}{\partial S_{i, k}}\right]\\
    & =  \eez \left[\nabla_{\rvx_i} \log \frac{q(\rvx_\theta;\theta_s)}{p(\rvx_\theta)}\cdot \rvz_k \right].
\end{split}
\end{equation*}
Hence, $\nabla_S \KL(q_\theta||p)_{i, k}$ is the mean of the product of the $i$-th element of $\gradx \log \frac{q(\rvx_\theta;\theta)}{p(\rvx_\theta)}$ and the $k$-th element of $\rvz$. We can write $ \nabla_S \KL(q_\theta||p)$ as
\begin{equation*}
\begin{split}
     \nabla_S \KL(q_\theta||p) &=  \mathbb{E}_{\rvz \sim \pz}\left[ \left(\gradx \log \frac{q(\rvx_\theta;\theta)}{p(\rvx_\theta)} \right)^T \cdot  \rvz \right]  \\
     &= \eez \left[ \left(\gradx \log \frac{q(\rvx_\theta;\theta)}{p(\rvx_\theta)}\right)^T\cdot  (\rvx_\theta-\mu) S^{-T}\right],  \quad \rvz =(\rvx_\theta-\mu) S^{-T}\\
      & = -\mathbb{E}_{\rvx \sim q_\theta}\left[\left(\gradx \log \frac{p(\rvx)}{ q(\rvx;\theta)} \right)^T(\rvx - \mu) S ^{-T}\right], \text{  by LOTUS}.
\end{split}
\end{equation*}
\\
\\
\\

\section{Riemannian Geometry} \label{appb1}
\subsection{Preliminaries}
\textbf{1. The space of non-singular matrices}:

The space of $\setS$ can be endowed with the Riemannian structure given a metric tensor $\gG$ induced by the Frobenius inner product. That is, for a point $S \in \setS$, we denote its tangent space as $\gT_S \setS$, then for $A, B \in \gT_S \setS$, metric tensor $\gG$ is given by
\begin{equation*}
    \langle A, B \rangle_{\gG} = \mathrm{tr}(A^TB).
\end{equation*}

Given a (smooth) functional $\gF:\setS \to \mathcal{R} $, 
the Riemannian gradient $\mathrm{grad}_{\gG} \gF(S) \in \gT_S \setS$ is defined as,
\begin{equation} \label{eq55}
   \forall X \in \gT_S \setS, \qquad \langle \mathrm{grad}_{\gG} \gF(S), X \rangle_{\gG} = \dd \gF_S(X),
\end{equation}
where $\dd \gF_S(X)$ is the differential of $\gF$ given by
\begin{equation*}
    \dd \gF_S(X) =\lim_{t \to 0} \frac{\gF(S + tX) - \gF(S)}{t} = \mathrm{tr}(\nabla_S \gF(S)^T X).
\end{equation*}
\begin{tcolorbox}[colback=NavyBlue!10,colframe=NavyBlue!15, boxrule=1.5mm, arc=5mm]
Therefore, 
$$
\langle \mathrm{grad}_{\gG} \gF(S), X \rangle_{\gG} = \mathrm{tr}(\nabla_S \gF(S)^T X) \Longrightarrow \mathrm{grad}_{\gG} \gF(S) = \nabla_S \gF(S),
$$
the Riemannian gradient on the manifold $(\setS, \gG)$ is the matrix derivative itself (Euclidean gradient). \\
\\
It is intuitive to think of the manifold $(\setS, \gG)$ as being analogous to the Euclidean space $\mathbb{R}^n$ of vectors since $\gG$ is a flat metric tensor.
\end{tcolorbox}
\textbf{2. Riemannian submersion}:

 Let $(\setC, \gQ)$ be another manifold with the metric tensor $\gQ$ and $\pi$ be a smooth map $\pi: \setS \to \setC$. If the differential map $\ddpi_S: \gT_S \setS \to \gT_{\pi(S)} \setC$ is surjective, the tangent space $\gT_S \setS$ can be decomposed into a vertical space $\gV_S$ and a horizontal space $\gH_S$,
 \begin{equation*}
     \gT_S \setS = \gV_S \oplus \gH_S.
 \end{equation*}
 The vertical space is the kernel of the differential map which comprises all elements that are mapped to zeros,
 \begin{equation*}
      \gV_S = \gK\ddpi_S = \{X| \ddpi_S(X)=0\},
 \end{equation*}
and the horizontal space is its orthogonal complement with respect to the metric tensor $\gG$, 
 \begin{equation*}
      \forall Y\in \gH_S \text{ and }  \forall X\in \gV_S, \quad \langle Y, X \rangle_\gG = 0.
 \end{equation*}
Geometrically, the kernel $\gK\ddpi_S$ represents the directions in which the mapping $\pi(S)$ is locally constant near the point $S\in \setS$.

We say the map $\pi: \setS \to \setC$ is a \textbf{Riemannian submersion}
if and only if for $S \in \setS$, the differential map $\ddpi_S$ is surjective and it maps the horizontal space $\gH_S$ to $\gT_{\pi(S)}\setC$ isometrically, i.e.,
    \begin{equation*}
                \langle \ddpi_S(X), \ddpi_S(Y)\rangle_{\gQ} = \langle X,Y \rangle_{\gG}, \quad X,Y \in \gH_S.
    \end{equation*}
    \\
\textbf{3. Properties of $\pi(S)=SS^T$}:

The differential of the map is given by $\ddpi_S(X) =  XS^T + SX^T, \quad X \in \gT_S\setS $. 

By the definition, the vertical space $\gV_S$ is given by the kernel $\gK\ddpi_S=\{X|XS^T+SX^T=0\}$. 

To find the horizontal space $\gH_S$, let $X \in \gV_S$ which indicates $XS^T$ is skew-symmetric, $\forall Y \in \gH_S$, $Y$ is orthogonal to $X$ by
\begin{equation*}
\begin{split}
    & \langle Y, X\rangle_\gG=\mathrm{tr}(Y^TX)=0\\
   & \Longrightarrow \mathrm{tr}(Y^TXS^TS^{-T})=0\Longrightarrow \mathrm{tr}(S^{-T}Y^TXS^T)=0\Longleftrightarrow YS^{-1}\text{ is symmetric.}
    \end{split}
\end{equation*}
This gives the horizontal space $\gH_S = \{Y|YS^{-1}\text{ is symmetric}\}$.  Note that for $Z \in \gT_S \setS$, only its horizontal component is mapped to $\gT_{\pi(S)} \setC$ because the vertical component is mapped to zero.
\\

\textbf{4. The space of positive-definite matrices}:

By Theorem 3 and Theorem 5 by \citep{bhatia2019bures}, for $\pi(S)=SS^T$ to qualify as a Riemannian submersion, there exists a unique metric tensor $\gQ$ on $\setC$, for $A=\pi(S)=SS^T \in \setC$, and $Y, Z \in \gT_A \setC $ given by the differential of the map $Y = \ddpi_S (HS)= HA + AH$ and $Z = \ddpi_S (KS)= KA + AK$, where $H$ and $K$ are symmetric matrices to ensure $HS$ and $KS$ lie within the horizontal space, the inner product of $\gQ$ must be given by
\begin{equation*}
    \langle Y,Z \rangle_{\gQ} = \langle HA + AH,KA + AK\rangle_{\gQ} = \langle HS,KS\rangle_{\gG} = \mathrm{tr}(KAH)
\end{equation*}

such that the induced distance function by this metric tensor $\gQ$ is the Bures distance $\mathcal{B}$, 
\begin{equation*}
    \forall A,B \in \setC, \qquad \mathcal{B}^2(A, B)=\Vert A^\frac{1}{2} -B^\frac{1}{2}U \Vert^2_F=\mathrm{tr}(A+B-2(A^{\frac{1}{2}}BA^{\frac{1}{2}})^{\frac{1}{2}}),
\end{equation*}
where $U= B^\frac{1}{2}A^\frac{1}{2}(A^{\frac{1}{2}}BA^{\frac{1}{2}})^{-\frac{1}{2}}$ is a unitary matrix by the polar decomposition of $A^\frac{1}{2}B^\frac{1}{2}$. $B^\frac{1}{2}U$ can be written as 
\begin{equation*}
    B^\frac{1}{2}U=T_{A\to B}  A^\frac{1}{2}, \quad T_{A\to B} = A^{-\frac{1}{2}}(A^{\frac{1}{2}}BA^{\frac{1}{2}})^{\frac{1}{2}}A^{-\frac{1}{2}},
\end{equation*}
where $T_{A\to B}: \sR^n \to \sR^n$ is the optimal transport map of moving mass from $\gN(0, A)$ to $\gN(0, B)$.

The geodesic connecting $A, B \in \setC$ under the metric tensor $\gQ$ is 
\begin{equation*}
    \Sigma_t = \big((1-t)I+ t T_{A\to B}\big)  A \big((1-t)I+ t T_{A\to B}\big), \quad t \in [0, 1],
\end{equation*}
where $\Sigma_0=A$ and $\Sigma_1=B$.

The Bures distance can be expressed by the length of another geodesic on $\setS$ connecting $A^\frac{1}{2}, B^\frac{1}{2}U \in \setS$ under the metric tensor $\gG$, 
\begin{equation*}
    S_t = \big((1-t)I+ t T_{A\to B}\big)  A^\frac{1}{2}, \quad t \in [0, 1],
\end{equation*}
where $S_0=A^\frac{1}{2}$ and $S_1=B^\frac{1}{2}U$.

\subsection{Proof of \autoref{hor}} \label{proof_p2}
Here we assume $q = \mathcal{N}(0, \Sigma)$ where the covariance matrix is $\Sigma = SS^T$,
and the gradient of the KL divergence w.r.t. $S$ is given by the \autoref{rule},
\begin{equation*}
     \nabla_S \KL(q_\theta||p)  = -\mathbb{E}_{\rvx \sim q_\theta}\left[\left(\gradx \log \frac{p(\rvx)}{ q(\rvx;\theta)} \right)^T \rvx S ^{-T}\right].
\end{equation*}
Using $\log q(\rvx;\theta)= - \rvx S^{-T}S^{-1} \rvx^T $,  $\gradx \log q(\rvx;\theta)= - \rvx S^{-T}S^{-1}  $ and $\gradx^2 \log q(\rvx;\theta)= -S^{-T}S^{-1}  $, we rewrite 
\begin{equation*}
\begin{split}
    -\mathbb{E}_{\rvx \sim q_\theta}\left[\left(\gradx \log \frac{p(\rvx)}{ q(\rvx;\theta)} \right)^T \rvx S ^{-T}\right] & =  \mathbb{E}_{\rvx \sim q_\theta}\left[\left(\gradx \log \frac{p(\rvx)}{ q(\rvx;\theta)} \right)^T \gradx 
 \log q(\rvx;\theta) \right]\cdot S \\
 & = \left[  \int \left(\gradx \log \frac{p(\rvx)}{ q(\rvx;\theta)} \right)^T\gradx q(\rvx;\theta)   \dd \rvx \right]\cdot S \\
 & = -\left[  \int \gradx^2 \log \frac{p(\rvx)}{ q(\rvx;\theta)}  q(\rvx;\theta)   \dd \rvx \right]\cdot S \\
 & =  -\eqx \left[ \gradx^2 \log p(\rvx) - \gradx^2 \log q(\rvx;\theta) \right] \cdot S \\
  & =  -\eqx \left[ \gradx^2 \log p(\rvx) \right] \cdot S -  S^{-T}
 \end{split}
\end{equation*}
$ \eqx \left[ \gradx^2 \log p(\rvx) \right]$ is the expectation of the Hessian matrix, hence it is symmetric, let it be $A$ such that we write the gradient of the KL divergence as
\begin{equation*}
    \nabla_S \KL(q_\theta||p) = -AS - S^{-T}.
\end{equation*}
The horizontal space is expressed by
\begin{equation*}
    \mathcal{H}_S = \{X| XS^{-1} \text{ is symmetric} \}.
\end{equation*}
Let $H = -A - S^{-T}S^{-1}$ such that $ \nabla_S \KL(q_\theta||p) = -AS - S^{-T} = HS$.
$A$ and $S^{-T}S^{-1}$ are symmetric matrices, hence $H$ is symmetric  $\Longrightarrow \nabla_S \KL(q_\theta||p) \in \mathcal{H}_S$.

\subsection{Proof of \autoref{lemma11}} \label{proof_l1}
Given two functionals: $\gF: \setS \to \mathcal{R}$ and $\gE: \setC \to \mathcal{R}$ satisfying
$$\gF(S) = \gE(\pi(S)).$$
For $X\in \gT_S \setS$,  we have the differential $\dd \gF_S(X)= \dd \gE_{\pi(S)}(\dd \pi_S(X))$ by chain rule. Now, \myeqref{eq55} rewrites as 
\begin{equation}
    \forall X \in \gT_S \setS, \qquad \langle \mathrm{grad}_{\gG} \gF(S), X \rangle_{\gG} = \dd \gE_{\pi(S)}(\dd \pi_S(X)),
\end{equation}
By the assumption $\mathrm{grad}_{\gG} \gF(S)$ is horizontal, so that $\mathrm{grad}_{\gG} \gF(S)$ is orthogonal to the vertical component of $X$. Hence, we have 
$$
\langle \mathrm{grad}_{\gG} \gF(S), X \rangle_{\gG} = \langle \mathrm{grad}_{\gG} \gF(S), X_{\gH} \rangle_{\gG},
$$
where $X_{\gH}$ represents the horizontal component of $X$.\\
\\
 Since $\pi$ is a Riemannian submersion which gives
 $
    \langle \ddpi_S(X), \ddpi_S(Y)\rangle_{\gQ} = \langle X,Y \rangle_{\gG}, \quad X,Y \in \gH_S$,
thus we have 
$$ \langle \ddpi_S(\mathrm{grad}_{\gG} \gF(S)), \ddpi_S(X)\rangle_{\gQ}=\langle \mathrm{grad}_{\gG} \gF(S), X_{\gH} \rangle_{\gG} =   \dd \gE_{\pi(S)}(\dd \pi_S(X))$$
By the definition of the Riemannian gradient 
$$\forall Y \in \gT_{\pi(S)}\setC, \quad \langle \mathrm{grad}_{\gQ}\gE(\pi(S)), Y\rangle_{\gQ} =  \dd \gE_{\pi(S)}(Y),$$
the Riemannian gradient on $\setC$ with the metric tensor $\gQ$ must be given by
$$
\mathrm{grad}_{\gQ}\gE(\pi(S)) = \ddpi_S(\mathrm{grad}_{\gG} \gF(S)).
$$

\subsection{Mapping A Curve Under Riemannian Submersion} \label{riemann}




Under the Riemannian submersion, a curve $\{ S_t\}_{t \geq0}$ in $\setS$ is a horizontal lift if it satisfies
\begin{equation}
    \frac{\dd S_t}{\dd t} = H(t) S_t,
\end{equation}
if $H(t)$ is a symmetric matrix. Obviously from \autoref{hor}, the Euclidean gradient flow $\{ S_t\}_{t \geq0}$ is horizontal with the symmetric matrix
\begin{equation}
    H(t) = \eqt \left[ \gradx^2 \log p(\rvx) \right]   +  S_t^{-T}S_t^{-1}.
\end{equation}
The corresponding curve in $\setC$ is 
\begin{equation}
\begin{split}
    \frac{\dd \Sigma_t}{\dd t} &= H(t) \Sigma_t + \Sigma_t H(t) \\
    &= 2I + \eqt \left[ \gradx^2 \log p(\rvx) \right] \Sigma_t +  \Sigma_t \eqt \left[ \gradx^2 \log p(\rvx) \right],
\end{split}
\end{equation}
this is equal to the Hessian form in Eq.~(\ref{hessian_form}).\\
\begin{tcolorbox}[colback=NavyBlue!10,colframe=NavyBlue!15, boxrule=1.5mm, arc=5mm]
The curve $\{ S_t\}_{t \geq0}$ is the horizontal lift of the curve $\{ \Sigma_t\}_{t \geq0}$. For any point $S_0$ in the fiber $\pi^{-1}(\Sigma_0)$, there is a unique curve $\{ S_t\}_{t \geq0}$ such that $\pi(S_0) = \Sigma_0$ and its image under the map $\pi \{ S_t\}_{t \geq0} = \{ \Sigma_t\}_{t \geq0}$. This means the Bures-Wasserstein gradient flow can be translated into the Euclidean gradient flow and if the initial point of the Euclidean gradient flow is given, then the flow curve is unique.
\end{tcolorbox}

Notice that the horizontal space $\gH_S = \{X| XS^{-1} \text{ is symmetric} \}$ indicates $\forall X \in \gH_S, XS^{-1}$ is symmetric. In another coordinate system, we can have $\dd \pi_S (X) = XS^{-1}$, this gives another form of Riemannian gradient,
\begin{equation}
    \mathrm{grad}\KL(q_\theta||p) =  \nabla_S \KL(q_\theta||p) S^{-1},
\end{equation}
which is the Bures-Wasserstein gradient of the KL divergence w.r.t. the covariance matrix studied by \citep{altschuler2021averaging, lambert2022variational, diao2023forward}

\textbf{Remark}. The Frobenius distance function $\Vert A - B \Vert_{\text{F}}^2=\text{tr}(AA^T+BB^T-2A^TB)$ is not the intrinsic Riemannian distance by the metric tensor $\gG$ because the space of non-singular matrices $\setS$ is not connected, i.e., $\forall A,B \in \setS$, the straight line segment $Z(t) = tA + (1-t)B, t\in[0, 1]$ may cross the set of singular matrices. However, the space $\setS$ can be separated into two connected subspaces: the set of positive-definite matrices $\setS_+$ and the set of negative-definite matrices $\setS_-$. Each can be equipped with the Frobenius distance given by the metric tensor $\gG$. If the curve $\{ \Sigma_t\}_{t \geq0}$ is smooth, its horizontal lift has to stay within the connected area, e.g., if the initialization $S_0 \in \setS_+$, the curve $\{ S_t\}_{t \geq0}$ stays within $\setS_+$. On the other hand, if we use the Monte Carlo method to evaluate gradients, there is always a small perturbation added to the matrix $S_t$ to avoid being singular matrices. Practically, we can ignore the disconnected property of $\setS$.

\newpage
\section{The Path-Derivative Gradient of $f$-Divergence}
\subsection{Proof of \autoref{first_var}}\label{le1}
Given the $f$-divergence as
\begin{equation*}
 \mathcal{F}_{f}(q)=\mathcal{D}_{f}(p||q) =  \int  f\left(\frac{p(\rvx)}{q(\rvx)} \right)q(\rvx) \dd \rvx.
\end{equation*}
Let $\phi \in \mathcal{P}(\mathbb{R}^n) $ be an arbitrary test function, the first variation  $\frac{\delta \gF_f}{\delta q}$ is given by
\begin{equation*}
\begin{split}
\int \frac{\delta \gF_f}{\delta q} (\rvx) \phi(\rvx) \mathrm{d}\rvx &= \lim_{\tau \to 0} \frac{\gF_f(q+\tau\phi)- \gF_f(q) }{\tau} \\
&=\frac{d}{d\tau} \gF_f(q+\tau\phi)\Big \vert_{\tau=0}\\
& = \frac{d}{d\tau} \int f\left(\frac{p}{q+\tau\phi}\right)(\rvx)\big(q(\rvx)+\tau\phi(\rvx)\big) \mathrm{d}\rvx\Big \vert_{\tau=0} \\
&=  \int  \left\{f\left(\frac{p}{q+\tau\phi}\right) (\rvx) \phi(\rvx)- f'\left(\frac{p}{q+\tau\phi}\right)(\rvx) \frac{p(\rvx) \phi(\rvx)}{q(\rvx)+\tau\phi(\rvx)} \right\} \mathrm{d}\rvx   \Big \vert_{\tau=0} \\
&=  \int  \left\{f\left(\frac{p}{q}\right) - f'\left(\frac{p}{q}\right) \frac{p}{q} \right\} (\rvx)\phi(\rvx) \mathrm{d}\rvx.
\end{split}
\end{equation*}
Thus,
\begin{equation*}
\frac{\delta \gF_f}{\delta q} = f(r) - rf'(r), \quad \text{where} \quad r = \frac{p}{q} .
\end{equation*}
This gives the Wasserstein gradient $\wgrad \gF_f(q) = \gradx \big[f(r) - rf'(r)\big]$.

\subsection{Proof of \autoref{p3}} \label{proof_p3} 
First, we write $f$-divergences 
$$
 \mathcal{D}_{f}(p||q_\theta) =\mathbb{E}_{\rvx \sim q_\theta} \big[ f(r(\rvx;\theta))  \big], \text{ where } r(\rvx;\theta) =\frac{p(\rvx)}{q(\rvx;\theta)},
$$
as 
\begin{equation}
\begin{split}
   \mathcal{D}_{f}(p||q_\theta)& =\eqx \big[ r f'(r) - r f'(r) + f(r) \big]\\
   &=\epx \big[f'\big(\ratiov\big)\big] - \eqx \big[ \ratiov f'\big( \ratiov\big) -f\big(\ratiov\big) \big].\label{dual_f_o}
    \end{split}
\end{equation}
Eq.~(\ref{dual_f_o}) is also a result from the dual representation of $f$-divergences \citep{nguyen2010estimating}.
We next apply $\dtheta$ to both sides of Eq.~(\ref{dual_f_o}),
\begin{equation}
\begin{split}
    \dtheta \mathcal{D}_{f}(p||q_\theta)& = \dtheta\mathbb{E}_{\rvx \sim p}\big[f'\big(\ratiov\big)\big] - \dtheta \mathbb{E}_{\rvx \sim q_\theta}\big[ \ratiov f'\big( \ratiov\big) -f\big(\ratiov\big) \big].\label{dual_f}\\ 
    \end{split} 
\end{equation}
\\
Notice that the reparameterization is given by $$\rvx_\theta = g(\rvz;\theta) \sim q(\rvx;\theta), \rvz \sim \pz.$$
The first term of the R.H.S. of Eq.~(\ref{dual_f}) can be written as 
\begin{equation}
\begin{split}
    \dtheta\mathbb{E}_{p(\rvx)}\big[f'\big(\ratiov\big)\big] &= \int p(\rvx) \dtheta f'\big(\ratiov\big) \dd \rvx\\
    &=\int p(\rvx) f''\big(\ratiov\big) \cdot \dtheta \ratiov \dd \rvx \\
    &=\int q(\rvx;\theta) \ratiov f''\big(\ratiov\big) \cdot \dtheta \ratiov \dd \rvx,  \text{  via the importance weight} \\
    &= \int \pz(\rvz) \ratio f''\big(\ratio\big)  \cdot \dtheta r(\rvx;\theta)\vert_{\rvx=\rvx_\theta}\dd \rvz , \text{  by LOTUS}\\ \label{first_term}
\end{split}
\end{equation}
The second term of the R.H.S. is 
\begin{equation}
\begin{split}
    &\dtheta \eqx\big[ \ratiov f'\big( \ratiov\big) -f\big(\ratiov\big) \big] \\
    & =  \dtheta\eez \big[ \ratio f'\big( \ratio\big) -f\big(\ratio\big) \big] , \text{  by LOTUS} \\
    & = \int \pz(\rvz) \dtheta \big[ \ratio f'\big( \ratio\big) -f\big(\ratio\big) \big]  \dd \rvz, \\
    & = \int \pz(\rvz)  \ratio f''\big( \ratio\big)  \cdot \dtheta r(\rvx_\theta;\theta)  \dd \rvz,\\
    & = \int \pz(\rvz)  \ratio f''\big( \ratio\big) \cdot \left[\gradx r(\rvx_\theta;\theta) \circ \dtheta \rvx_\theta + \dtheta r(\rvx;\theta)\vert_{\rvx=\rvx_\theta}\right]\dd \rvz, \text{  by Eq.~(\ref{61})}.
    \label{second_term}
\end{split}
\end{equation}

Eq.~(\ref{first_term}) - Eq.~(\ref{second_term}), we have 
\begin{equation}
\label{path-df}
\begin{split}
    \dtheta \mathcal{D}_{f}(p||q_\theta)& = -\eez \left[ r(\rvx_\theta;\theta) f''\big(r(\rvx_\theta;\theta)\big) \cdot \gradx r(\rvx_\theta;\theta) \circ \dtheta \rvx_\theta \right]\\
     & = -\eez \left[   h'\big( r(\rvx_\theta; \theta_{s})\big) \cdot \dtheta r(\rvx_\theta;\theta_s)\right], \text{  by Eq.~(\ref{62})} \\
    & = -\eez \left[ \dtheta  h\big( r(\rvx_\theta; \theta_{s})\big) \right],
    \end{split}
\end{equation}
where $h(r)=rf'(r) - f(r)$.\\
\\
We summarize some typical $f$-divergences and their associated $h$ functions in \autoref{expilcit form}. It can be seen that for all $\alpha$-divergences where $h'(r) = r^{\alpha-1}$, if the target $p(\rvx)$ is unnormalized, we have
\begin{equation}\label{constand_aff}
 h'\left(\frac{p(\rvx)}{q(\rvx)}\right) \propto h'\left(\frac{p_{\mathrm{true}}(\rvx)}{q(\rvx)}\right),   
\end{equation}

where $p_{\mathrm{true}}(\rvx) = p(\rvx) / \int p(\rvx) \dd \rvx$. Hence, according to Eq.~(\ref{path-df}), the normalizing constant only affects the scales of the path-derivative gradient, which can be folded into the learning rate.
\begin{table}[h]
\caption{$f$-divergences and associated $h$ functions.}
\label{expilcit form}
\begin{center}
\begin{tabular}{llll}
\multicolumn{1}{c}{}  &\multicolumn{1}{c}{$f(r)$}  & $h(r)=rf'(r) - f(r)$ & $h'(r)$\\
\toprule 
Reverse KL     ($\alpha=0$)         &$-\log r$              & $\log r -1$        &$\frac{1}{r}$\\
Forward KL  ($\alpha=1$)           &$r\log r$              & $r$                &$1$\\
$\chi^2$    ($\alpha=2$)        &$(r-1)^2$              & $r^2-1$            &$2r$\\
Hellinger   ($\alpha=0.5$)            &$(\sqrt{r}-1)^2$       & $\sqrt{r}-1$       &$\frac{1}{2\sqrt{r}}$\\
$\alpha$-divergence  ($\alpha \neq 0, 1$)      &$\frac{r^\alpha-\alpha r - (1-\alpha)}{\alpha(\alpha-1)}$   &$\frac{r^\alpha -1}{ \alpha}$ &$r^{\alpha -1}$\\
\bottomrule
\end{tabular}
\end{center}
\end{table}

\newpage
\section{Experiments}
Codes are available on \url{https://github.com/YiMX/Bridging-the-gap-between-VI-and-WGF}.
\subsection{The Illustrative Example on Gaussians} \label{exp1}
\begin{figure*}[h]

     \centering

\includegraphics[width=0.45\textwidth]{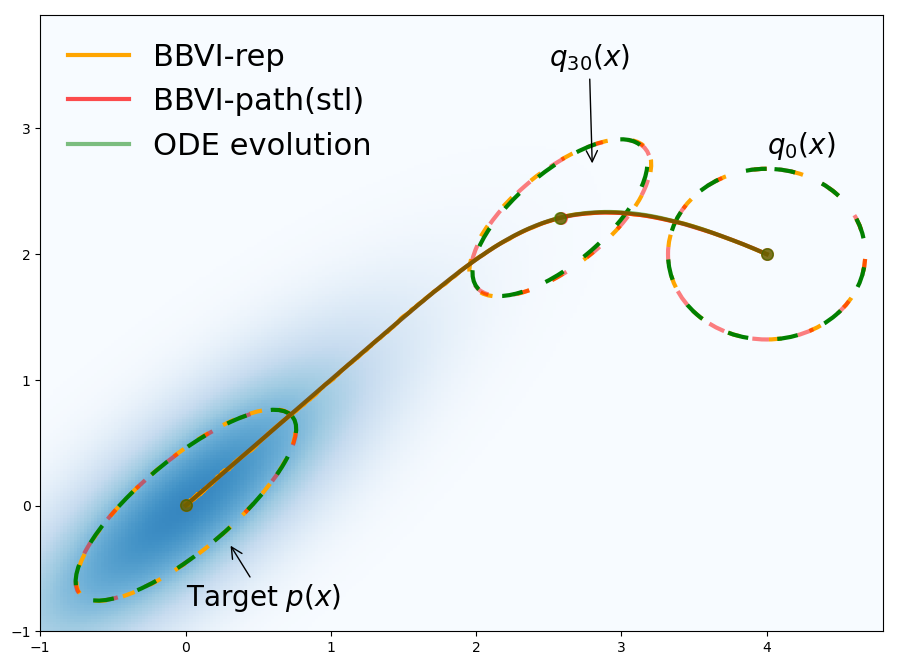}
  \includegraphics[width=0.45\textwidth]{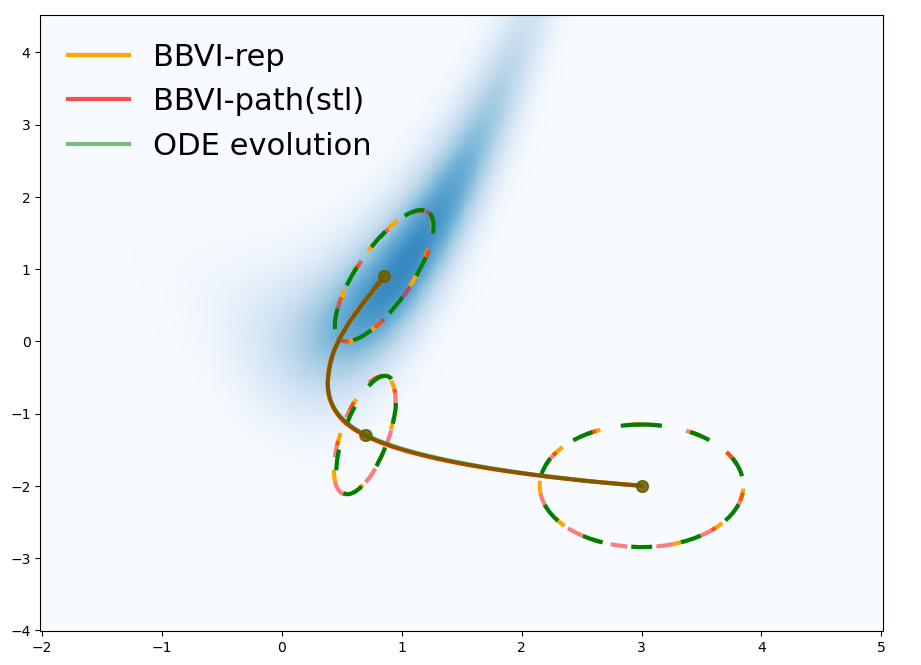}   
     \caption{Mean and covariance evolution of Gaussians. The sample size for the Monte Carlo gradient is 100. }
     \label{app1}
\end{figure*}

In Section \ref{illus1}, the target distribution $p(\rvx)$ is a Gaussian $\gN(\mu, \Sigma)$ with $\mu=(0.0, 0.0)$ and $\Sigma= ((0.8, 0.4),(0.4, 0.8))$. The initial variational distribution is Gaussian with $\mu=(4.0, 2.0)$ and identity covariance matrix $I$. In \autoref{illus}, we use 5 particles to evaluate the Monte Carlo gradients for each algorithm and the learning rate is set to be 0.01. In \autoref{app1}, the sample size of the Monte Carlo gradient increases to 100, we can observe that the variance of BBVI-rep becomes smaller and all three algorithms still generate the same visible evolution. The target density in the right figure in \autoref{app1} follows the Rosenbrock density function, 
\begin{equation}
    p(\rvx) \propto \exp\left\{-a(\rvx_1 - \mu)^2-b(\rvx_2 -\rvx_1^2)^2\right\}, \quad \rvx = (\rvx_1, \rvx_2), \label{91}
\end{equation}
where $a=1.0, b=1.0, \mu=1.0$.

\subsection{On The Path-Derivative Gradient of $f$-Divergences} \label{final_exp}
The path-derivative gradient of $f$-divergences is given by
\begin{equation}
    \dtheta \mathcal{D}_{f}(p||q_\theta) =  - \eez\left[ \dtheta h\big( r(\rvx_\theta; \theta_{s})\big) \right],
\end{equation}
which defines a surrogate Monte Carlo objective $$L(\theta) = - \eez\left[h\big( r(\rvx_\theta; \theta_{s})\big) \right].$$ To this end, we can evaluate this surrogate Monte Carlo objective and differentiate it to get the unbiased gradient estimator of an $f$-diveregence such that we can perform BBVI, as shown in  \autoref{path-vi}.
\begin{algorithm}
\caption{BBVI with the path-derivative gradient of $f$-divergence} \label{path-vi}
\begin{algorithmic}
\Require unnomarlized density $p(\rvx)$, variational family $q(\rvx;\theta)$, the associated $h(r)=rf'(r)-f(r)$ and learning rate $\tau$.
\While{not converged}
 \State 1. Sample $N$ particles $\{\rvz^i\}_{i=1\cdots N} \sim \pz$ and reparameterizing  $\rvx_\theta^i = g(\rvz^i;\theta)$.
 \State 2. Compute $r(\rvx_\theta^i; \theta_s) = p(\rvx_\theta^i)/q(\rvx_\theta^i;\theta_s)$, where stop gradient operator is applied to $\theta$.
 \State 3. Compute $L(\theta)=  -\frac{1}{N}\sum_i h(r(\rvx_\theta^i, \theta_s))$.
 \State 4. $\theta \leftarrow \theta - \tau \dtheta L(\theta)$ via back-propagation.
 \EndWhile
\end{algorithmic}
\end{algorithm}

\subsubsection{Toy Example}
\begin{figure}[h]
\vspace{-0.0cm}
     \centering
     \begin{subfigure}{0.23\textwidth}
         \centering

         \includegraphics[width=\textwidth]{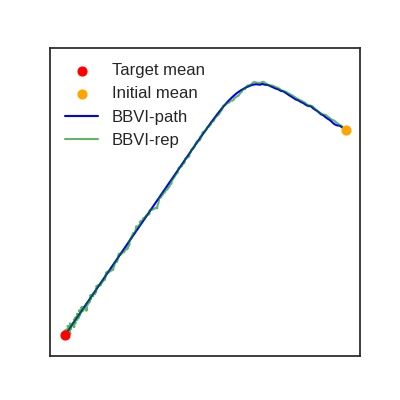}
         \vspace{-8mm}
     \end{subfigure}
     \begin{subfigure}{0.24\textwidth}
         \centering
         \includegraphics[width=\textwidth]{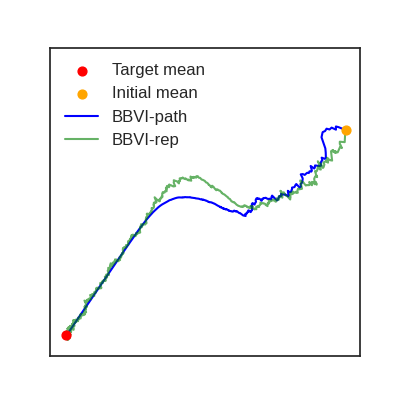}
         \vspace{-8mm}
     \end{subfigure}
      \begin{subfigure}{0.24\textwidth}
         \centering
         \includegraphics[width=\textwidth]{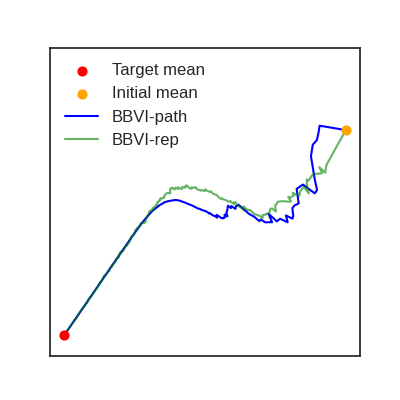}
         \vspace{-8mm}
     \end{subfigure}
      \begin{subfigure}{0.24\textwidth}
         \centering
         \includegraphics[width=\textwidth]{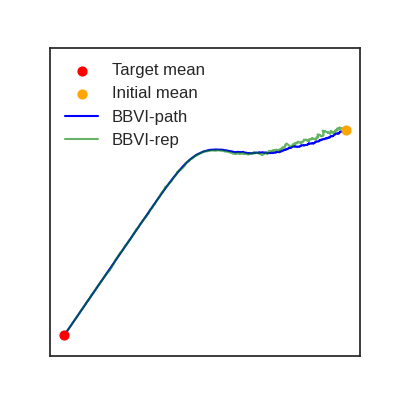}
         \vspace{-8mm}
     \end{subfigure}
     \begin{subfigure}{0.24\textwidth}
         \centering
         \includegraphics[width=\textwidth]{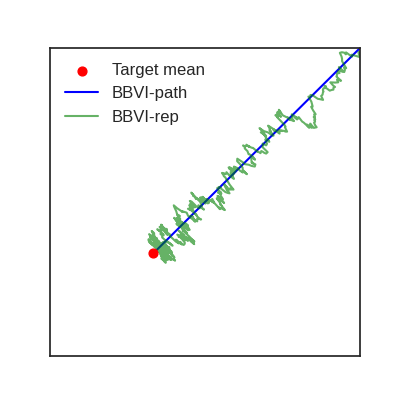}
          \caption{Reverse KL}
         \vspace{-2mm}
     \end{subfigure}
     \begin{subfigure}{0.24\textwidth}
         \centering
         \includegraphics[width=\textwidth]{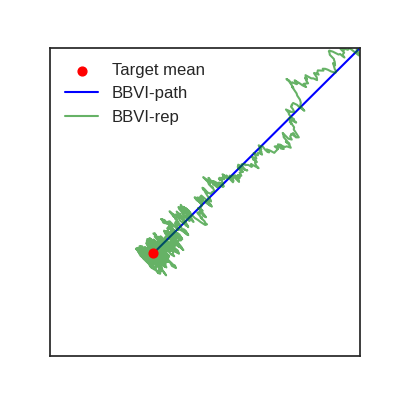}
         \caption{Forward KL}
         \vspace{-2mm}
     \end{subfigure}
      \begin{subfigure}{0.24\textwidth}
         \centering
         \includegraphics[width=\textwidth]{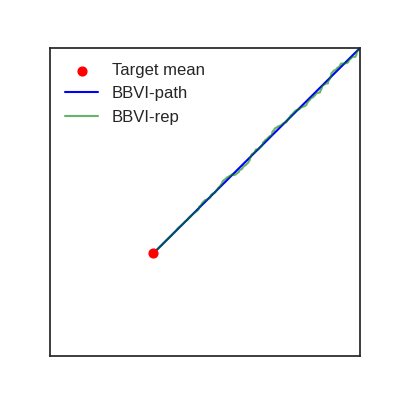}
         \caption{$\chi^2$}
         \vspace{-2mm}
     \end{subfigure}
      \begin{subfigure}{0.24\textwidth}
         \centering
         \includegraphics[width=\textwidth]{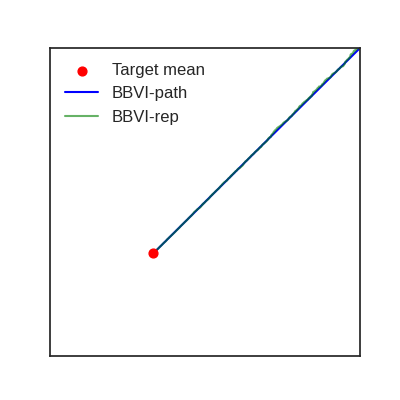}
         \caption{Hellinger}
         \vspace{-2mm}
     \end{subfigure}
     \caption{Trajectories of means of Gaussian variational distributions. The bottom figures are the plots zoomed in around the target mean.}
     \label{trac}
\end{figure}

In this section, we illustrate \autoref{path-vi} using different $f$-divergences. The target distribution is an unnormalized 2D Gaussian $\gN(\mu, \Sigma)$ with $\mu=(0.0, 0.0)$ and $\Sigma= ((0.5, 0.3),(0.3, 0.5))$ and the variational distribution is also a 2D Gaussian initialized at $\mu=(1.0, 0.5)$ and $\Sigma= ((1.0, 0.0),(0.0, 1.0))$. We plot the trajectories of the means of Gaussian variational distributions in \autoref{trac}. For comparison, we also plot the trajectories obtained via BBVI using the reparameterization gradient (BBVI-rep). To allow for computing the reparameterization gradient, we borrow the ground truth normalized target distribution's density. In \autoref{trac}, we can observe that different $f$-divergences produce different trajectories of Gaussian means in Euclidean space, this corresponds to distilling different gradient flows (curves of marginal probabilities) in Wasserstein space. We also observe that the trajectories of BBVI-path exactly evolve to the target mean under all $f$-divergences, whereas the trajectories of BBVI-rep fluctuate around the target mean under reverse KL divergence and forward KL divergence. This phenomenon corresponds to Eq.~(\ref{83}) where the variance of the path-derivative gradient diminishes if the variational distribution well approximates the target distribution such that the Wasserstein gradient becomes zero, also known as "sticking the landing" \citep{roeder2017sticking}. This fluctuation of BBVI-rep does not happen under $\chi^2$ divergence and Hellinger divergence. 

\subsection{Bayesian Logistic Regression}
In this part, we implement the path-derivative gradient for Bayesian logistic regression using UCI dataset \cite{asuncion2007uci}. For comparison, the baseline is the standard VI method--reverse KL with reparameterization trick and "CHIVI" method \citep{dieng2017variational}. Note that the reparameterization gradient is intractable due to the normalizing constant, except for the reverse KL. The variational family is diagonal Gaussian. In this experiment, we use the same trick as \cite{dieng2017variational} to adjust the density ratio per iteration in \autoref{path-vi} to enhance the numerical stability,
\begin{equation}
    r(\rvx_\theta^i;\theta_s) = \exp\big[\log r(\rvx_\theta^i;\theta_s) - \max_i \log r(\rvx_\theta^i;\theta_s) \big],
\end{equation}
since the constant $\max_i \log r(\rvx_\theta^i;\theta_s)$ only affects the scale of the path-derivative gradient according to Eq.~(\ref{constand_aff}).

Below is the test set accuracy with standard deviation, calculated with 32 posterior samples. In \autoref{bayesian}, we found VI with the path-derivative gradient generalizes well to different datasets compared to the standard VI method and CHIVI.
\begin{table}[h]
\caption{Test accuracy. Higher is better.}
\label{bayesian}
\begin{center}
\resizebox{\columnwidth}{!}{%
\begin{tabular}{lllllll}
Dataset&	RKL (rep)&	
 CHIVI&	RKL (path)&	FKL (path)	&
 $\chi^2$ (path)	&Hellinger (path)\\
\toprule 
Heart &$0.871\pm0.017$	&$0.786\pm0.080$	&$0.872\pm0.024$	&$0.815\pm0.069$	&$0.792\pm0.066	$&$0.828\pm0.055$\\
Ionos      &$0.783\pm 0.022$&$0.670\pm 0.094$&$0.782\pm 0.022$&$0.665\pm 0.090$&$0.664\pm 0.094$&$0.664\pm 0.103$ \\
Wine         &$0.720\pm 0.015$	&$0.694\pm 0.041$	&$0.720\pm 0.015$	&$0.693\pm 0.046$	&$0.692\pm 0.047$	&$0.703\pm 0.035$\\
Pima          &$0.775\pm 0.016$	&$0.731\pm0.037$	&$0.776\pm 0.013$	&$0.726\pm 0.035$	&$0.733\pm 0.037$	&$0.748\pm 0.029$\\
\bottomrule
\end{tabular}}
\end{center}
\end{table}

\subsection{Extension to Gaussian Mixture Models}
In this section, we discuss how to apply the path-derivative gradient to update the parameters of Gaussian mixture variational families. Using Gaussian mixture models (GMMs) enriches the flexibility of the approximation. A GMM comprises $K$ individual Gaussian distributions,  we denote $m_k$ as the weight and $\theta_k$ as the parameter for the $k$-th Gaussian component. The probability density function of GMM is 
\begin{equation}
    q(\rvx;\theta, m) = \sum^K_{k=1} m_k q_k(\rvx;\theta_k), \quad \sum^K_{k=1} m_k=1.
\end{equation}
The reparameterization path of the sample $\rvx_{\theta,m}\sim q(\rvx;\theta, m)$ from GMMs is not directly differentiable since it requires discrete sampling from a categorical distribution to determine in which component the sample is generated.
We should notice the surrogate Monte Carlo objective $L(\theta, m)$ for GMMs can be decomposed via conditional sampling as
\begin{equation}
   L(\theta, m) =-\mathbb{E}_{\rvx \sim q(\rvx;\theta, m)} \big[h(r(\rvx;\theta_s, m_s)\big]= -  \sum_{k=1}^K m_k \eez \big[h(r(\rvx_{\theta_k};\theta_s, m_s)\big],
\end{equation}
where $\rvx_{\theta_k} =g(\rvz; \theta_k)$ is sampled from the $k$-th component distribution $q_k(\rvx;\theta_k)$. With the help of the stop gradient operator, the surrogate Monte Carlo objective disentangles the interaction of the parameters of GMMs such that the gradients for each $\theta_k$ and $m_k$ depend only on samples from the $k$-th component. We give a summary of the distilled Wasserstein gradient flows with GMM variational families in \autoref{gmm}.

\begin{algorithm}
\caption{Variational Inference with Gaussian Mixture Variational Families}\label{gmm}
\begin{algorithmic}
\Require unomarlized density $p(\rvx)$, variational distribution $q(\rvx;\theta, m)=\sum^K_{k=1} m_k q_k(\rvx;\theta_k)$, the associated function $h(r)=rf'(r)-f(r)$ and learning rate $\tau$.
\While{not converge}
 \For {$k=1\cdots K$}
 \State {1. Sample $N$ particles $\{\rvz^i\}_{i=1\cdots N} \sim \pz$ and reparameterizing  $\rvx_{\theta_k}^i = g(\rvz^i;\theta_k)$}.
  \State {2. Compute $r_i = p(\rvx_{\theta_k}^i)/q(\rvx_{\theta_k}^i;\theta_{s}, m_{s})$, where the stop gradient operator is applied.}
  \State {3. Compute $\ell_k =  \frac{1}{N}\sum_i h(r_i)$}.
 \EndFor

 \State {4. Compute $ L(\theta, m) = -  \sum_{k=1}^K m_k \ell_k$}.
 \State {5. $(\theta, m) \leftarrow (\theta, m) - \tau \nabla_{(\theta, m)} L(\theta, m)$ and via back-propagation}.
 \EndWhile
\end{algorithmic}
\end{algorithm}
We show how  \autoref{gmm} performs on approximating Rosenbrock density in Eq.~(\ref{91}) (banana distributions). The number of components is set to $5$ and we add Softmax activations to ensure the sum of weights is equal to 1, the approximated variational GMMs are shown in \autoref{funcs} with contour plots of their density functions. 

\subsubsection{Approximating 1D Gaussian Mixture Distributions}
\begin{figure}[h!]
     \makebox[20pt]{\raisebox{40pt}{\rotatebox[origin=]{90}{$K=1$}}}%
     \begin{subfigure}[b]{0.233\textwidth}
         \centering
         \includegraphics[width=\textwidth]{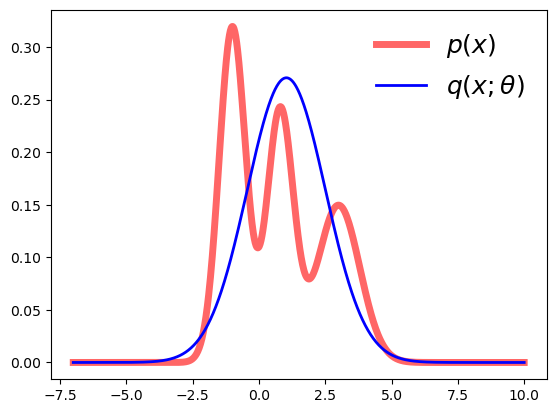}
     \end{subfigure}
     \begin{subfigure}[b]{0.233\textwidth}
         \centering
         \includegraphics[width=\textwidth]{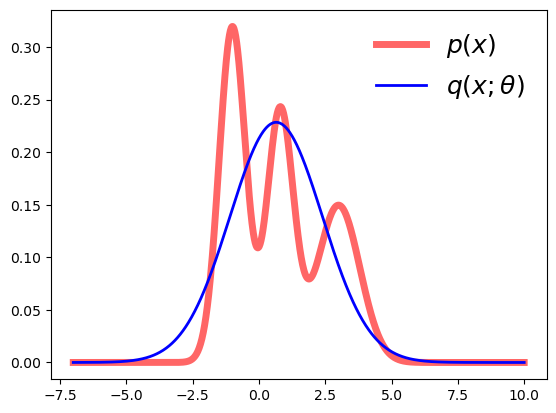}
     \end{subfigure}
     \begin{subfigure}[b]{0.233\textwidth}
         \centering
         \includegraphics[width=\textwidth]{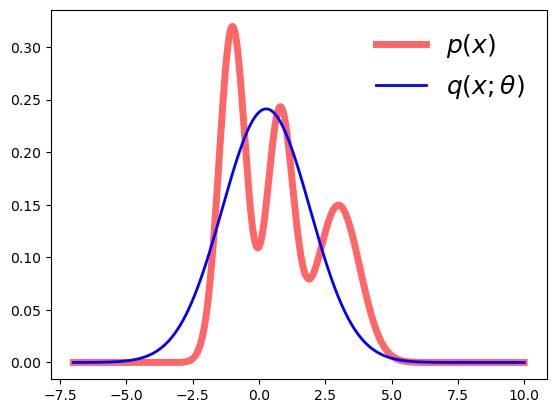}
     \end{subfigure}
     \begin{subfigure}[b]{0.233\textwidth}
         \centering
         \includegraphics[width=\textwidth]{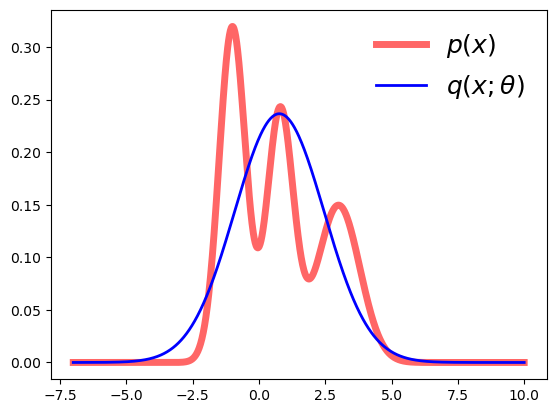}
     \end{subfigure}
     \makebox[20pt]{\raisebox{40pt}{\rotatebox[origin=c]{90}{$K=2$}}}%
\begin{subfigure}[b]{0.233\textwidth}
         \centering
         \includegraphics[width=\textwidth]{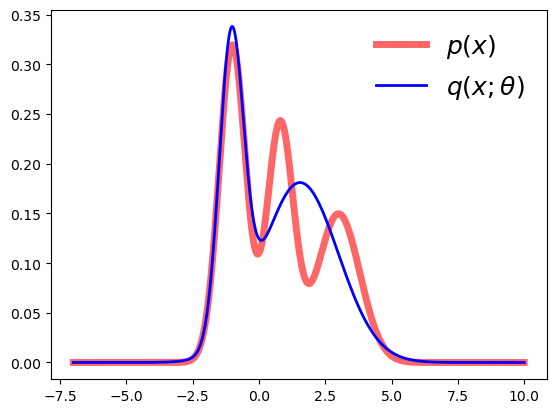}
     \end{subfigure}
     \begin{subfigure}[b]{0.233\textwidth}
         \centering
         \includegraphics[width=\textwidth]{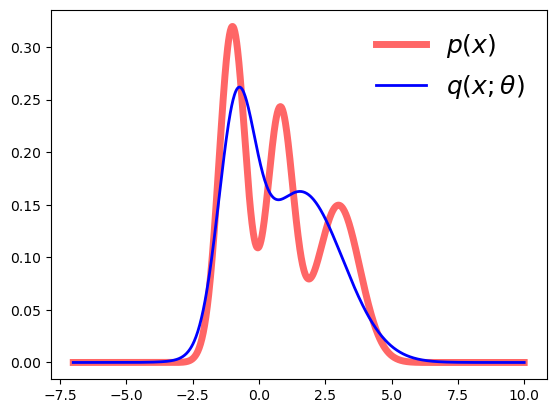}
     \end{subfigure}
     \begin{subfigure}[b]{0.233\textwidth}
         \centering
         \includegraphics[width=\textwidth]{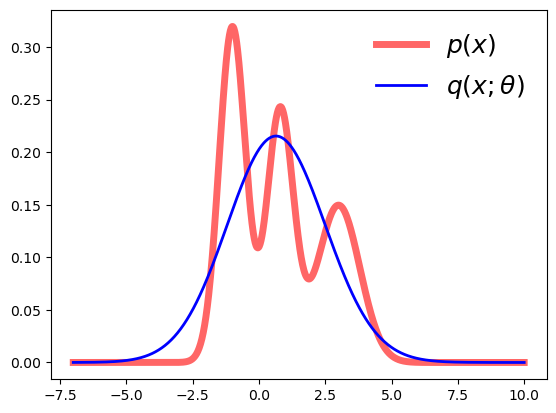}         
     \end{subfigure}
     \begin{subfigure}[b]{0.233\textwidth}
         \centering
         \includegraphics[width=\textwidth]{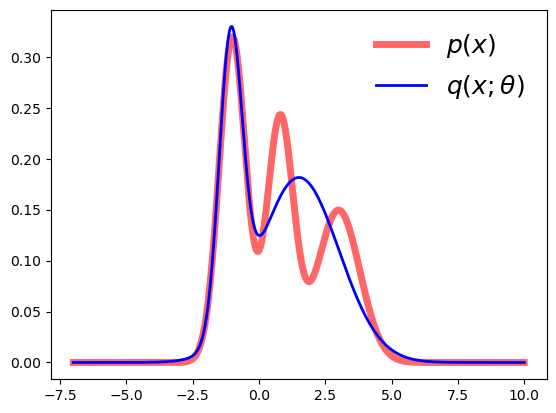}
     \end{subfigure}
     \makebox[20pt]{\raisebox{40pt}{\rotatebox[origin=c]{90}{$K=3$}}}%
\begin{subfigure}[b]{0.233\textwidth}
         \centering
         \includegraphics[width=\textwidth]{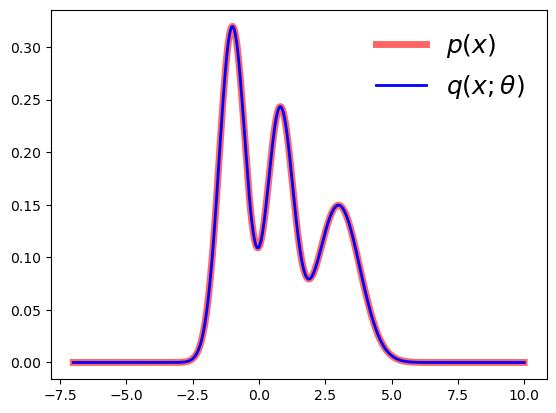}
     \end{subfigure}
     \begin{subfigure}[b]{0.233\textwidth}
         \centering
         \includegraphics[width=\textwidth]{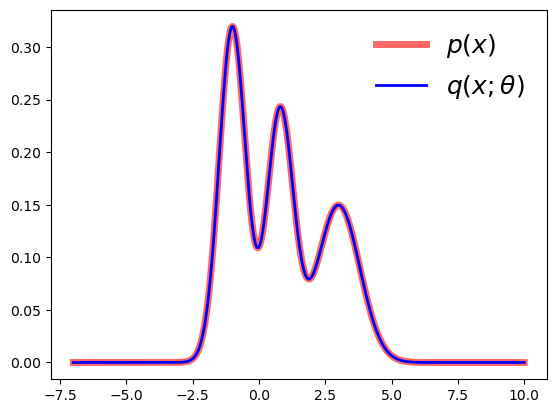}
     \end{subfigure}
     \begin{subfigure}[b]{0.233\textwidth}
         \centering
         \includegraphics[width=\textwidth]{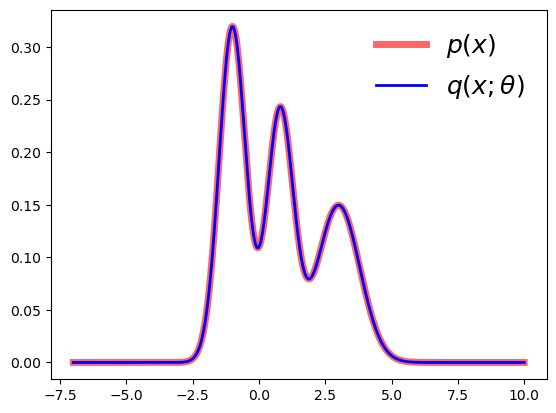}
     \end{subfigure}
     \begin{subfigure}[b]{0.233\textwidth}
         \centering
         \includegraphics[width=\textwidth]{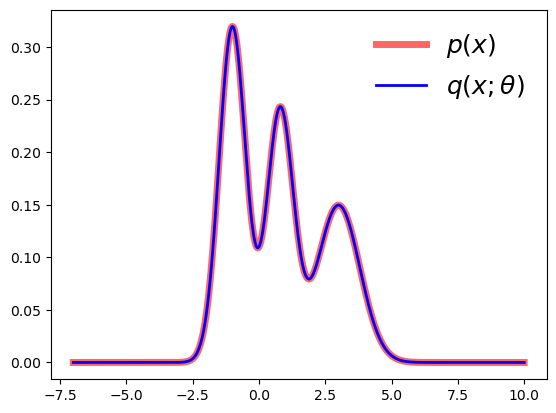}
     \end{subfigure}
     \makebox[20pt]{\raisebox{40pt}{\rotatebox[origin=c]{90}{$K=4$}}}%
  \begin{subfigure}[b]{0.233\textwidth}
         \centering
         \includegraphics[width=\textwidth]{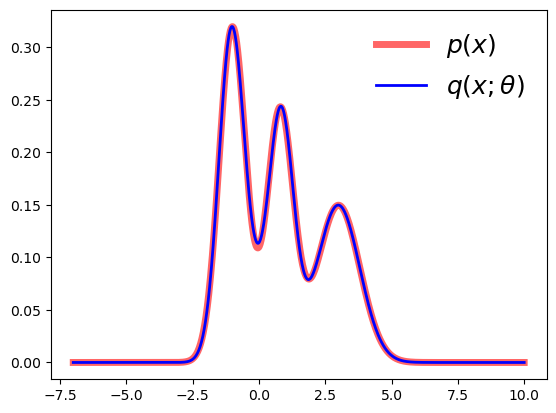}
          \caption{Reverse KL}
     \end{subfigure}
     \begin{subfigure}[b]{0.233\textwidth}
         \centering
         \includegraphics[width=\textwidth]{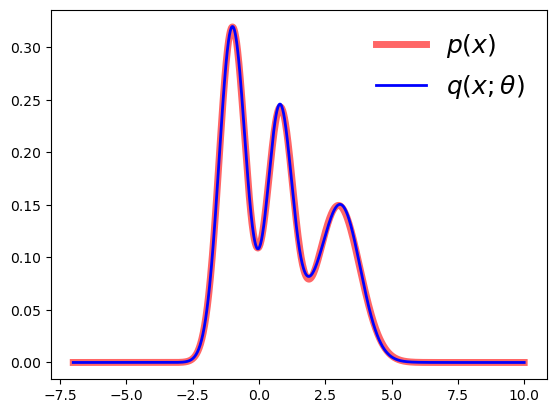}
         \caption{Forward KL}
     \end{subfigure}
     \begin{subfigure}[b]{0.233\textwidth}
         \centering
         \includegraphics[width=\textwidth]{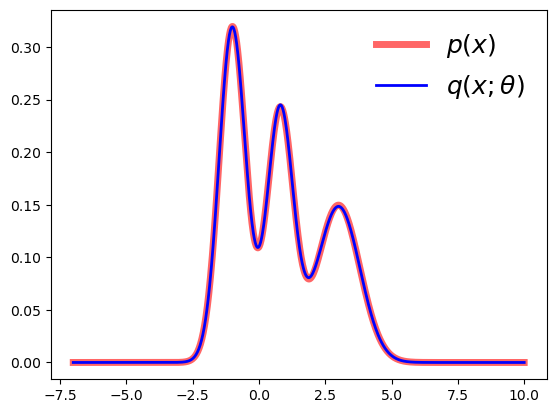}
         \caption{$\chi^2$}
     \end{subfigure}
     \begin{subfigure}[b]{0.233\textwidth}
         \centering
         \includegraphics[width=\textwidth]{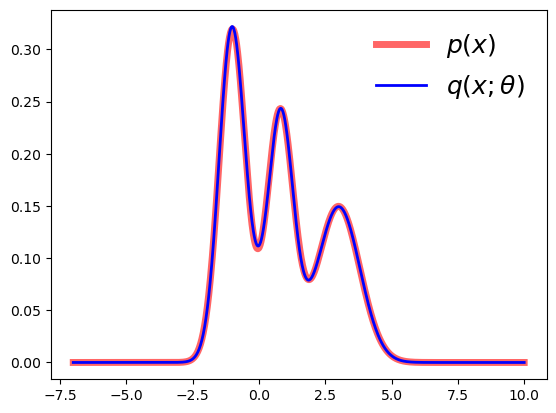}
         \caption{Hellinger}
     \end{subfigure}
     \caption{Approximating a target Gaussian mixture distribution via variational GMMs with different numbers of components. }
     \label{1dgmm}
\end{figure}

The target distribution is a 3-mode Gaussian mixture distribution with density function with a normalizing constant, $$p(\rvx) \propto 0.4\mathcal{N}(-1.0, 0.25) + 0.3\mathcal{N}(0.8, 0.25) + 0.3\mathcal{N}(3.0, 0.64).$$ The variational distribution has the density function, $$q(\rvx;\theta, m) = \sum^K_{k=1} m_k q_k(\rvx;\theta_k), \quad \sum^K_{k=1} m_k=1,$$
where each $q_k(\rvx;\theta_k)$ is a Gaussian distribution. We implement \autoref{gmm} to approximate this target under $f$-divergences via Gaussian mixture variational families that have different numbers of components $K$. The density plots of the target distribution and variational distributions are reported in Figure \ref{1dgmm}. We can observe that with $K$ increases, the resulting approximations are more accurate.

\end{document}